\documentclass{article}
\usepackage{log_2024}	
\usepackage{dsbda-style}
\usepackage{booktabs}						%
\usepackage{multirow}						%
\usepackage{amsfonts}						%
\usepackage{graphicx}						%
\usepackage{duckuments}						%
\usepackage{placeins}
\usepackage{adjustbox}
\usepackage{subfigure}
\usepackage{adjustbox}
\usepackage{graphicx} %
\usepackage{subcaption} %
\usepackage{url}

\usepackage{easyReview}
\setreviewsoff
\usepackage{comment}
\usepackage[numbers,compress,sort]{natbib}	%

\title{Edge-Splitting MLP: Node Classification on Homophilic and Heterophilic Graphs without Message Passing}

\author[M.Kohn et al.]{%
Matthias Kohn\\
University of Ulm\\
Germany\\
\email{matthias.kohn@uni-ulm.de}\And
Marcel Hoffmann\\
University of Ulm\\
Germany\\
\email{marcel.hoffmann@uni-ulm.de}\And
Ansgar Scherp\\
University of Ulm\\
Germany\\
\email{ansgar.scherp@uni-ulm.de}
}
 
\begin{document}

\maketitle

\begin{abstract}
Message Passing Neural Networks (MPNNs) have demonstrated remarkable success in node classification on homophilic graphs.
It has been shown that they do not solely rely on homophily but on neighborhood distributions of nodes, \ie consistency of the neighborhood label distribution within the same class. 
MLP-based models do not use message passing, \eg Graph-MLP incorporates the neighborhood in a separate loss function.
These models are faster and more robust to edge noise.
Graph-MLP maps adjacent nodes closer in the embedding space but is unaware of the neighborhood pattern of the labels, \ie relies solely on homophily.
Edge Splitting GNN (ES-GNN) is a model specialized for heterophilic graphs and splits the edges into task-relevant and task-irrelevant, respectively.
To mitigate the limitations of Graph-MLP on heterophilic graphs, we propose ES-MLP that combines Graph-MLP with an edge-splitting mechanism from ES-GNN.
It incorporates the edge splitting into the loss of Graph-MLP to learn two separate adjacency matrices based on relevant and irrelevant feature pairs.
Our experiments on seven datasets with six baselines show that ES-MLP is on par with homophilic and heterophilic models on all datasets without using edges during inference.
We show that ES-MLP is robust to multiple types of edge noise during inference and that its inference time is two to five times faster than that of commonly used MPNNs.
The source code is available at \url {https://github.com/MatthiasKohn/ES-MLP}
\end{abstract}

\section{Introduction}
In social networks, people tend to be friends with people with similar interests, and in citation graphs, papers tend to cite papers of the same subject~\cite{BeyondHomophilyInGNN}.
The property of such graphs that adjacent nodes are likely to share the same class is denoted as homophily.
Homophily is the underlying assumption of many Message Passing Neural Networks (MPNNs)~\cite{HeteroGNNSurvey}.
However, in the real world, many graphs are heterophilic, \ie adjacent nodes tend to have different labels. 
For example, fraudsters in online transactions are more likely to build connections with customers instead of other fraudsters~\cite{HeteroGNNSurvey} or different types of amino acids interact in protein structures~\cite{BeyondHomophilyInGNN}.
Several works claim that classical MPNNs have been designed under the homophily assumption and are unsuitable for heterophilic graphs~\cite{BeyondHomophilyInGNN}.
Hence, new Graph Neural Networks (GNNs) have been developed, \eg Edge Splitting GNN (ES-GNN)~\cite{Edge-SplittingGNN}, that are more capable of handling heterophilic graphs through higher-order neighborhood aggregation or GNN architecture refinement methods~\cite{HeteroGNNSurvey, ACM-GNN}.
Recent work showed that MPNNs can perform well on some heterophilic graphs if the graph shows a uniform neighborhood pattern~\cite{IsHomophilyANecessity}.
However, these findings do not apply to multilayer perception (MLP) models for graphs as they do not rely on message passing.
Graph-MLP~\cite{GraphMLP:abs-2106-04051} is a popular GNN without messaging passing. 
It is based on an MLP and a neighborhood contrastive loss to combine the edge and feature information in a single embedding during training.
Hence, Graph-MLP does not require edges during inference.
The neighborhood contrastive loss of Graph-MLP uses only single edges during training, \ie considers only a pair of connected vertices at a time.
Thus, Graph-MLP is unaware of the neighborhood patterns that are required to perform well on heterophilic graphs~\cite{IsHomophilyANecessity}.

We propose Edge-Splitting MLP (ES-MLP), which combines the strength of Graph-MLP and ES-GNN.
ES-GNN splits the original graph edges into two exclusive sets to mitigate the problem of MPNNs on heterophilic graphs.
Graph-MLP does not use message passing, unlike conventional MPNNs.
This results in higher computational efficiency and robustness to edge noise during inference.
With ES-MLP, we propose a model for heterophilic graphs that does not need access to the adjacency matrix during inference time as well but is able to reduce the influence of harmful edges.
In contrast to ES-GNN, our ES-MLP is able to add new edges since the computations are based on powers of the adjacency matrix.
Our model is robust against edge noise, \ie differences in the neighborhood distribution of the training and test dataset, and has low computation time during inference.
In our experiments, we evaluate the performance of ES-MLP against five baseline GNN models on seven real-world datasets ranging from low homophily to high homophily and a synthetic dataset generated with the Contextual Stochastic Block Model (CSBM)~\cite{GraphAttentionRetrospective(CSBM)}.
In summary, our contributions are:
\begin{itemize}
    \item We propose ES-MLP which combines the benefits of Graph-MLP and ES-GNN to handle both homophilic and heterophilic graphs without message passing.

    \item We evaluate the effectiveness of ES-MLP on seven benchmark datasets and one synthetic dataset and show that it is on par with all baselines in classification and inference time performance.

    \item We show that ES-MLP is robust to multiple types of edge noise during inference time.

\end{itemize}

\section{Related Work}
\label{sec:relatedwork}

\paragraph{GNNs under Homophily}
~\citet{GraphConvNet} proposed the Graph Convolutional Network (GCN) for node classification, which performs message passing by aggregating the neighborhood features to obtain the node embedding.
\citet{SymplifyingGCN} simplified it by removing the nonlinearities between GCN layers to collapse the resulting function into a single linear transformation.
~\citet{GraphAttentionNetwork} introduced Graph Attention Network (GAT), an attention-based architecture that computes the feature representation of every node, by weighting edges differently based on an attention mechanism.
GraphSAGE~\cite{GraphSage} separates the weight of a node from its neighbors, which makes it strong on heterophilic graphs.
~\citet{GraphMLP:abs-2106-04051} proposed Graph-MLP as a pure MLP-based framework. 
It avoids explicit message passing by using the node features as input of an MLP and calculating a separate contrastive loss function based on the powers of the adjacency matrix.

 \paragraph{GNNs under Heterophily}
Most of the GNN models are designed under the assumption of homophily and perform low on heterophilic graphs~\cite{BeyondHomophilyInGNN}.~\citet{PerformanceDiscrepanciesLocalHomphily} and \citet{PAC-Bayesian} demonstrated that node classification performance degrades when the local homophily of a node deviates from the global homophily.
Existing GNNs for heterophilic graphs can mainly be categorized into two types~\cite{HeteroGNNSurvey}, aggregating higher-order neighborhoods and propagating weighted messages.
MixHop~\cite{MixHop} aggregates messages from multi-hop neighbors.
H$_2$GCN~\cite{BeyondHomophilyInGNN} is designed with a propagation mechanism, which automatically changes the propagation and aggregation process according to homophily or heterophily between node pairs.
The second type is passing weighted messages between heterophilic neighbors~\cite{GGCN, GPR-GNN, ACM-GNN, NCC_Metric_TA-GCN}.
GGCN uses cosine similarity to send signed neighbor features under certain constraints of relative node degrees.
GPR-GNN learns weights that can be positive or negative for feature propagation.
TA-GCN~\cite{NCC_Metric_TA-GCN} estimates the neighborhood patterns of the graph data and learns guided by this metric an augmented graph topology.
\citet{Label-WiseGraphConvolution} proposed Label-Wise GCN, which summarizes the neighborhood information of every node by label-wise aggregation to capture the useful heterophilic context.
\add{\citet{GBK-GNN} proposed GBK-GNN, which employs a bi-kernel feature transformation and a selection gate mechanism, where one kernel captures homophilic node pairs and the other heterophilic node pairs.}
~\citet{EdgeDirectionalyForHeteroGraphs} introduced Directed GNN (Dir-GNN), which treats the graph as directed and performs separate aggregations of the incoming and outgoing edges. 
Edge Splitting GNN~\cite{Edge-SplittingGNN} introduces an edge-splitting layer that disentangles edges into two exclusive sets, allowing the GNN to focus on the most relevant edges for the classification task. 
These two channels are aggregated separately and the prediction is made only on the task-relevant feature channel.
LINKX~\cite{LINKX} is an MLP-based model that processes the adjacency matrix and node features separately and then concatenates them as a simple baseline for heterophilic graphs.
In contrast to ES-MLP, LINKX requires edges during inference time.

\paragraph{Homophily Measures}
There are several measures of homophily for a graph.
~\citet{BeyondHomophilyInGNN} proposed edge homophily ratio $h_{edge}$, which measures the proportion of edges in a graph that connects nodes within the same class.
Another measure is the node homophily $h_{node}$ \cite{NodeHomoRatio}.
For each node, it measures the ratio of adjacent nodes with the same labels and all adjacent nodes. $h_{node}$ is the average over all nodes.
Based on the edge homophily measure,~\citet{ClassInsEdgeHomoRatio} proposed a class insensitive edge homophily ratio $h_{class}$.
This measure is normalized by the number and size of the classes, which makes graphs with different numbers and sizes of classes comparable. 
It takes the edge homophily measure~\cite{BeyondHomophilyInGNN} and subtracts the general proportion of nodes in the same class to all nodes to mitigate the class-imbalance problem.
Adjusted homophily $h_{adj}$~\cite{AdjustedHomophily} is the edge homophily adjusted for the expected number of edges connecting nodes with the same class, considering the number of classes and the distribution of node degrees among them.
We use the adjusted homophily as reference measure since it is comparable across different datasets with varying numbers of classes and class size balance.
The formulas of the homophily measures are provided in Appendix \ref{appendix: homophily measures}.

\section{Edge-Splitting MLP}
\label{sec:methods}
In this section, we propose our new model Edge-Splitting MLP (ES-MLP), which combines elements of Graph-MLP~\cite{GraphMLP:abs-2106-04051} and ES-GNN~\cite{Edge-SplittingGNN}. 
We assume a transductive setting for node classification, \ie test nodes are already present during training.

The key feature of ES-GNN is an additional edge-splitting (ES) layer to partition the network topology to disentangle the task-relevant and irrelevant node features. 
We denote task-relevant feature pairs that benefit from a connection through an edge.
In contrast to task-irrelevant features that are harmful to the task if connected by an edge.
The ES layer splits node features to distinguish the task-relevant and irrelevant connections among nodes.
The features are then aggregated separately along these connections to produce disentangled representations.
ES-GNN incorporates an Irrelevant Consistency Regularization (ICR) loss, which minimizes the distance of heterophilic neighbors.

Graph-MLP is an MLP-based model without message passing. Instead, it uses a neighborhood contrastive loss to bridge the gap between GNN and MLP by utilizing the adjacency information during training. 
For each node, the $r$-hop neighbors are regarded as positive samples and the other nodes as negative samples.

\begin{figure*}
    \centering
    \includegraphics[width=1.00\textwidth]{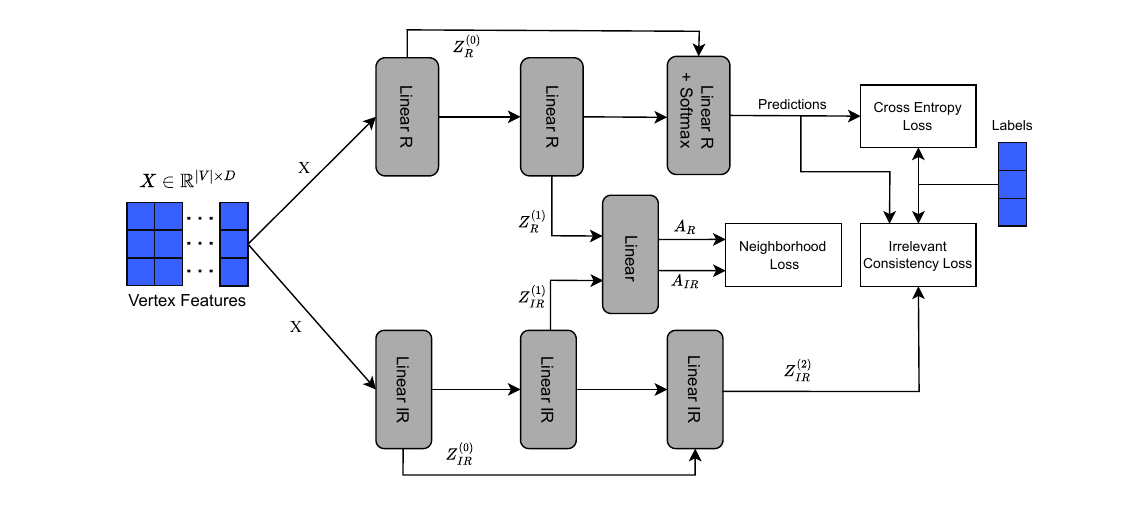}
    \caption{The pipeline of ES-MLP. Node features are first projected into two different subspaces $R$ and $IR$. A separate forward pass is performed to obtain embeddings $Z_R$ and $Z_{IR}$. Both embeddings are then used for edge splitting to divide the original graph edges into two sets and for the neighborhood contrastive loss. The task-relevant representation $Z_R$ is then used for the prediction task and task-irrelevant representation $Z_{IR}$ is used to calculate the Irrelevant Consistency Loss.}
    \label{fig:method_pipeline}
\end{figure*}

\paragraph{Edge-Splitting MLP}
 ES-MLP generalizes MLP to both, homophilic and heterophilic graph data.
 To this end, we integrate the neighborhood contrastive loss from Graph-MLP into ES-GNN to aggregate features on task-relevant edges without using the adjacency matrix during inference time. An overview of ES-MLP is given in Figure~\ref{fig:method_pipeline}. 

Let $G = \{V, E\}$ denote a graph, where $V$ and $E$ are the sets of nodes and edges, respectively. 
We project the feature matrix $X \in \mathbb{R} ^ {|V| \times d}$ into two different subspaces $Z_s$ with $s \in \{R, IR\}$ to obtain two separate representations, where $|V|$ is the number of nodes and $d$ the feature dimension. 
$Z_R$ represents task-relevant $(R)$ and $Z_{IR}$ task-irrelevant $(IR)$ embeddings.
The projection is defined by $Z_s^{(0)} = \sigma(XW_s+ b_s)$, where $W_s$ and $b_s$ are learnable parameters for each channel $s \in \{R,IR\}$. 
$R$ is the task-relevant channel and $IR$ is the task-irrelevant channel, and $\sigma$ is a nonlinear activation function.

The edge-splitting mechanism separates the graph edges into two exclusive sets of edges, which are represented by $A_R, A_{IR} \in \mathbb{R}^{|V| \times |V|}$, with splitting coefficients $A_{R(i,j)} \in [0,1]$ and $ A_{IR(i,j)} \in [0,1] $.
These matrices are then used for the neighborhood contrastive loss.
We process each channel $R$ and $IR$ seperately without using any message passing by $Z_s^{(k+1)} = \epsilon_s Z_s^{(0)} + (1 - \epsilon_s)Z_s^{(k)} W_s$,
with skip connections and weighting hyperparameter $\epsilon_s$.
We use a linear layer for the final classification.

The splitting coefficients $A_{R(i,j)} $ and $ A_{IR(i,j)}$ are exclusive, \ie $A_{R(i,j)} + A_{IR(i,j)} = A_{(i,j)} = 1$. Therefore, we parameterize the residual between $A_{R(i,j)}$ and $A_{IR(i,j)}$, and linear equations: 

\begin{displaymath}
    A_{R(i,j)} - A_{IR(i,j)} = \alpha_{(i,j)}, \quad
    A_{R(i,j)} + A_{IR(i,j)} = 1
\end{displaymath}

This leads to two exclusive sets of edges, which are used for the neighborhood contrastive loss: 

\begin{displaymath}
    A_{R(i,j)} = \frac{1 + \alpha_{(i,j)}}{2}, \hspace{0.2cm} A_{IR(i,j)} = \frac{1 - \alpha_{(i,j)}}{2},
\end{displaymath}
with $\alpha_{(i,j)} \in (-1,1)$.
The weights $\alpha_{(i,j)}$ influence both channels
and are computed by:

\begin{displaymath}
    \alpha_{(i,j)} = tanh(FF[Z_{R[i,:]}^{(K)} \oplus Z_{IR[i,:]}^{(K)} \oplus Z_{R[j,:]}^{(K)} \oplus Z_{IR[j,:]}^{(K)}]^T),
\end{displaymath}

where we concatenate the embeddings of both representations, $R$ and $IR$ of two connected nodes $v_i$ and $v_j$. 
These are then passed into a linear layer $FF$, followed by a $tanh$ activation function, to obtain values within $(-1,1)$.
It ensures the exclusiveness of our splitting coefficients.

 \paragraph{Neighborhood Contrastive Loss}

To incorporate the edge information, we use the Neighborhood Contrastive Loss from~\citet{GraphMLP:abs-2106-04051}.
This loss enables our model to learn graph structure without explicit message passing.
Since this loss is designed under the assumption that connected nodes are similar to each other, we calculate it on the task-relevant $R$ and task-irrelevant $IR$ embedding spaces separately to generalize the loss for heterophilic graphs.
The neighborhood contrastive loss for node $v_i$ can be formulated as:
\begin{displaymath}
    l_{i_s} = - log \frac{\sum_{j=1}^{|V|} 1_{[j \neq i]} \hat{\gamma}_{s(i,j)} exp(sim(Z_{s[i,:]}, Z_{s[j,:]})/\tau)} {\sum_{k=1}^{|V|} 1_{[k \neq i]} exp(sim(Z_{s[i,:]}, Z_{s[k,:]} )/\tau)} ,
\end{displaymath}

where $sim$ denotes the cosine similarity, $\tau$ a temperature parameter, $s \in \{R,IR\}$ and $\gamma_{s(i,j)} $ is calculated based on the $r^{th}$ power of the splitting adjacency matrices: $\gamma_{s(i,j)} = A_{s(i,j)}^r$ with $s \in \{R,IR\}$ and $\gamma_{s(i,j)}$ if node $v_i$ is in the $r$-hop neighborhood of node $v_j$.

Instead of simply minimizing the distance between nodes in the $r$-hop neighborhood, the loss minimizes the distance of node embeddings for edges that are relevant for the task according to feature embeddings.
Simultaneously, the task-irrelevant edges are learned based on the task-irrelevant part of the feature vectors.
Since taking the $r^{th}$ power of the splitting adjacency can lead to small splitting coefficients, we renormalize the $ \gamma_{s(i,j)}$ to ensure that $\gamma_{R(i,j)}  + \gamma_{IR(i,j)} = 1$.
The values of $\gamma_{s(i,j)} $ are non-zero if and only if node $j$ is in the $r$-hop neighborhood of node $v_i$~\cite{GraphMLP:abs-2106-04051}:
$$
\hat{\gamma}_{s(i,j)}=\begin{cases}
     \gamma_{s(i,j)}/ (\gamma_{R(i,j)} + \gamma_{IR(i,j)}), & \text{$j$ is in the $r$-hop neighborhood of $v_i$}\\
     0, & \text{ $j$ is not in the $r$-hop neighborhood of $v_i$}
\end{cases}
$$

The final neighborhood contrastive loss is computed by averaging over all nodes,    ${\mathcal{L}_{NC} = \frac{1}{|V|} \sum_{i=1}^{|V|} (l_{i_R} + l_{i_{IR}})}$.

Given that $Z_{IR}$ is designed to represent the least relevant information for the node classification task, an Irrelevant Consistency Regularization (ICR) is added to improve the consistency of irrelevant information in $Z_{IR}$.
The ICR loss is formulated as: 
 \begin{displaymath}
     \mathcal{L}_{ICR} = \sum_{(i, j) \in E} (1- \hat{y_i}^T \hat{y_j}) \lVert Z_{IR_i} - Z_{IR_j}\rVert_2,
 \end{displaymath}

 where $\hat{y_i}^, \hat{y_j}$ are the predicted probability vector of nodes $i$ and $ j$, and $Z_{IR_i}$, $ Z_{IR_j}$ denoting the irrelevant embedding of nodes $i$ and $j$.
 The ICR loss minimizes the distance of two adjacent nodes if they do not share the same label.

 Finally, the total loss is the sum of the ICR $\mathcal{L}_{ICR}$, neighborhood contrastive loss $\mathcal{L}_{NC}$, and the cross entropy loss $\mathcal{L}_{CE}$ :

\begin{displaymath}
 \mathcal{L}_{final} = \mathcal{L}_{CE} + \alpha_{NC} \cdot \mathcal{L}_{NC} + \beta_{ICR} \cdot \mathcal{L}_{ICR}, 
\end{displaymath}
 
where $\alpha$ is the neighborhood contrastive loss weight and $\beta$ the ICR loss weight.

\paragraph{Complexity Analysis}
The computational complexity for one training step of ES-MLP is $\mathcal{O}(dh^{l-2}|C|)$ for a forward pass through the network and another $\mathcal{O}(r|V|^3))$ to compute the $r$-th power of the adjacency matrix for the loss function, where $h$ is the hidden dimension, $|C|$ the number of classes, the $d$ feature dimension, and $l$ the number of layers.
This is more than for regular MPNNs, \eg GCN, which has a computational complexity of $\mathcal{O}(|E|dh^{l-2}|C|)$~\cite{GraphConvNet}.
However, ES-MLP has a computational complexity of only $\mathcal{O}(dh^{l-2}|C)|$ during inference, \ie has lower computational complexity than MPNNs.

In this section, we proposed ES-MLP, which splits the original graph edges into two exclusive sets of edges to aggregate features on task-relevant and irrelevant edges. 
This edge-splitting mechanism is incorporated in a neighborhood contrastive loss to bypass the use of message passing.
Model predictions are performed on task-relevant embeddings and an irrelevant consistency loss is used to suppress information in the task-relevant embedding.

\section{Experimental Apparatus}
\label{sec:experimentalapparatus}
\subsection{Datasets}
\label{sec:datasets}
\paragraph{Real World Datasets}
For our experiments, we use seven datasets, three with a homophily score over $0.5$ and four heterophilic datasets with homophily under $0.5$.
We measure homophily by the adjusted homophily ratio~\cite{AdjustedHomophily} since it can compare datasets with different numbers of classes.
Table~\ref{tab:DatasetsStatistics} shows the properties of each dataset.

\begin{table}[!ht]
        \caption{Statistics of the real-world datasets. 
        $|V|$ denotes the number of nodes, $|E|$ the number of edges, $d$ the feature dimension, $|C|$ the number of classes, $h_{adj}$ the adjusted homophily measure, and $h_{edge}$ the edge homophily.}
     \centering
    \begin{tabular}{l  c  c  c  c  c c}
    \toprule
      \textbf{Dataset}   & $|V|$ & $|E|$ & $d$ & $|C|$ & $h_{adj}$ & $h_{edge}$\\
         \hline
      Cora     & 2,708 & 5,278 & 1,433 & 7 & 0.77 & 0.81\\
      CiteSeer & 3,327 & 4,552 & 3,703 & 6 & 0.67 & 0.74\\
      PubMed   & 19,717 & 44,324 & 500 & 3 & 0.69 & 0.80\\
      \hline
      Actor & 7,600 & 30,019 & 932 & 5 & 0.00 & 0.22\\
      Roman & 22,662 & 32,927 & 300 & 18 & -0.05 & 0.05 \\
      Amazon & 24,492 & 93,050 & 300 & 5 & 0.14 & 0.38\\
      Minesweeper & 10,000 & 39,402 & 7 & 2 & 0.01 & 0.68\\
      \bottomrule
    \end{tabular}
    \label{tab:DatasetsStatistics}
\end{table}

The homophilic datasets are the three standard citation graphs Cora, CiteSeer, and PubMed~\cite{CitationNetworkDatasets}. 
In these datasets, each node represents a document and edges correspond to citations between them. 
For the heterophilic datasets, we use Actor~\cite{NodeHomoRatio}, Amazon-ratings (Amazon), Roman-empire (Roman), and Minesweeper \cite{NewHeteroData}. 
Actor represents actors where the edges describe co-occurrences in the same Wikipedia article.
In the Roman dataset, each node corresponds to one (non-unique) word in the Roman Wikipedia article.
The Amazon dataset is a co-purchase graph.
Minesweeper is based on the popular Minesweeper game and consists of a $100 \times 100$ grid.
Each node represents a cell.
The task is to classify each cell as a bomb or not a bomb.
Node features are one-hot-encoded numbers of adjacent bombs.
Note that in this dataset the features of a node are independent from its class.
A detailed explanation of the datasets can be found in Appendix~\ref{ExtendetDSDescription}.

\paragraph{Synthetic Dataset}
We use the Contextual Stochastic Block model (CSBM)~\cite{GraphAttentionRetrospective(CSBM)} to generate random graphs with controlled levels of homophily. 
A CSBM is defined as $X, A \sim CSBM(|V|, p, q, \mu, \sigma^2)$, where the edges are modeled by two random Bernoulli distributions $p$ and $q$.
If two nodes $v_i$, $v_j$ belong to the same class, the probability that an edge $e_{i,j}$ between these nodes is present is distributed with $e_{i,j}$ $\sim Ber(p), p \in [0,1]$. 
If the nodes $v_i$, $v_j$ belong to different classes, the probability that an edge is present is distributed with $e_{i,j}$ $\sim Ber(q), q \in [0,1]$. 
For each node $v_i$, we assign the $d$-dimensional feature vector $x_i \sim \mathcal{N}((2y_i - 1)\mu, \sigma^2\mathbf{I})$, where $y_i \in \{0,1\}$ is the class of node $v_i$, $\mu \in \mathbb{R}^{d}$, $\sigma \in \mathbb{R}$, and $\mathbf{I} \in \{0,1\}^{d \times d}$ is the identity matrix.
This model allows us to control the ratios of the inter and intra-class edges, \ie the homophily of the graph.

\subsection{Procedure}
\label{sec:procedure}

We compare our ES-MLP to six baseline models, MLP, GCN, Graph-MLP, ES-GNN, GraphSAGE, and \add{LINKX}. 
We train and evaluate the models on seven real-world and one synthetic datasets.
In addition, we perform a robustness analysis, where we add edge noise to the test graph.
For Minesweeper, we use AUROC since it is a binary classification task~\cite{NewHeteroData}.

\paragraph{Real-world Benchmarks}
For the homophilic graphs Cora, CiteSeer, and PubMed, we create $10$ random splits with $20$ nodes per class following~\citet{Pitfalls}.
For the heterophilic dataset Actor, we use the split from~\citet{NodeHomoRatio}.
It consists of 10 random splits with $48\%$ train nodes, $32\%$ validation nodes, and $20\%$ test nodes.
For Amazon, Roman, and Minesweeper, we adopt the setting from~\citet{NewHeteroData}.
We repeat each experiment $10$ times and compute the test accuracy on the node classification task.

\paragraph{Synthetic Benchmarks}
For the synthetic data, we generate a graph using a CSBM for multiple combinations of the parameters $p$ and $q$ from $0$ to $1$ with steps of $0.2$.
We fix the number of nodes to $5,000$  and use two balanced classes.
In total, we have 36 experiments, each is run 5 times. 
For the CSBM's number of features per node $d$, mean $\mu$, and standard deviation $\sigma$, we use a similar setting as \citet{GraphAttentionRetrospective(CSBM)}, \ie we set $d = n / \log^2(n)$ features per node, $\sigma = 0.20$ and $\mu = 10\sigma \sqrt{\log n^2}/2\sqrt{2d}$. 
We split the synthetic graphs into a $60\%$ train, $20\%$ validation, and $20\%$ test nodes.
The overlap in node feature distribution between the classes leads to an inevitable error probability of $0.25$. 
The actual error may be lower due to the additional information from the edges.

\paragraph{Robustness Analysis}

We investigate the robustness towards edge noise by deviating homophily levels during inference time.
We introduce two types of edge noise, which we increase gradually.
We adopt the edge noise addition algorithm provided by~\citet{IsHomophilyANecessity}.
The algorithm adds a fixed number of edges to the graph according to a chosen distribution.
It uniformly samples a node $v_i$ with label $y_i$.
Then a label $\Tilde{c}$ is sampled from a neighborhood class distribution $D_{y_i}$.
From the set of nodes $V_{\Tilde{c}}$ with class $\Tilde{c}$, we uniformly sample a node $v_j$.
Finally, a new edge $(v_i,v_j)$ is added to the graph.
We set the distributions $D_{\Tilde{c}}$ to be uniform to simulate uniform edge noise.
We also experiment with setting $D_{\Tilde{c}}$ to a categorical distribution to simulate a shift in neighborhood classes.
For the categorical noise, we adopt a circulate matrix-like design to make any two classes from $C$ have different distributions, see Appendix~\ref{appendix: distribution details} for the detailed distribution per class.
We perform this experiment for Cora and Amazon.

\paragraph{Inference Time}
We measure the inference time of the models in two settings, with the full graph as model input and only the test nodes as model input.
We aim to demonstrate that MLP-based models can be provided with only the test graph as input and do not decrease in performance.
This results in a performance gain in terms of inference time, an advantage, that is not possible with MPNNs, \eg GCN, which is expected to lose performance, when only test nodes are given as model input.

\paragraph{Hyperparameter Optimization}
\label{sec:hyperparameteroptimization}

We optimized the learning rate, hidden dimension, weight decay, and dropout for GCN, GraphSAGE, LINKX, and MLP. 
For GraphMLP, ES-GNN, and ES-MLP, we additionally tuned the loss weights $\alpha$ for neighborhood contrastive loss and $\beta$ for the irrelevant consistency loss and the used power of the adjacency matrix $r$.
Details on the search space and procedure, as well as the final hyperparameters, can be found in Appendix~\ref{hyperparameter_optim}.

\section{Results}
\label{sec:results}
\paragraph{Real-world Datasets}
Table~\ref{tab:results_real_world} shows the results for the real-world datasets. 
ES-MLP consistently shows competitive performance across all three homophilic datasets, where it keeps up with the best MPNN-based models.
On Cora and PubMed, ES-MLP outperforms the homophilic MLP-based model Graph-MLP.
On the heterophilic datasets Actor, Roman, and Amazon, ES-MLP outperforms all baselines.
GraphSAGE is the best model on Minesweeper.

\begin{table*}[!ht]
    \centering
    \adjustbox{max width=\textwidth}{
        \begin{tabular}{  c  c  c  c  c  c  c|  c}
        \toprule
        &  MLP & GCN & Graph-MLP &  ES-GNN & GraphSAGE & LINKX & ES-MLP (ours)  \\
        \hline
        Cora     & $76.95_{1.00}$ & $\mathbf{88.46_{0.83}}$  & $86.64_{1.14}$  & $87.30_{0.43}$ & $88.26_{0.50}$ & $83.15_{0.59}$ & $88.15_{1.85}$ \\
        CiteSeer & $72.10_{1.12}$ & $77.41_{0.95}$ & $\mathbf{77.79_{0.10}}$ & $74.27_{1.50} $& $76.54_{0.73}$ & $73.23_{0.85}$& $75.67_{0.92}$ \\
        PubMed   & $87.49_{0.90}$ & $\mathbf{89.63_{0.79}}$ & $87.06_2.41$   & $88.81_{0.49}$ & $89.60_{0.41}$ & $87.47_{0.29}$ & $87.56_{1.23}$\\
        \hline
        Actor          & $35.81_{0.62}$ & $29.24_{0.47}$ & $36.03_{0.98}$  & $38.91_{0.45}$ & $32.24_{0.76}$ & $33.92_{1.11}$ & $\mathbf{39.73_{0.37}}$\\
        Roman   & $60.50_{0.88}$ & $41.40_{1.58}$ & $64.94_{0.25}$  & $60.41_{1.90}$ & $62.47_{1.90}$ & $65.40_{0.37}$& $\mathbf{65.44_{0.92}}$\\
        Amazon & $44.05_{0.54}$ & $46.27_{0.67}$ & $37.07_{0.80}$   & $46.53_{0.34}$ & $44.83_{1.16}$ & $39.25_{0.51}$ & $\mathbf{47.8  5_{1.23}}$\\
        Minesweeper    & $50.54_{0.49}$ & $71.44_{0.74}$ & $50.99_{0.35}$ & $68.23_{1.10}$ & $\mathbf{88.90_{2.37}}$  & $51.61_{1.4}$&  $ 50.87_{2.03}$\\
        \bottomrule
        \end{tabular}}
    \caption{Test accuracies and standard deviation in percent (\%) on the real-world datasets averaged over $10$ runs for each experiment. The best score for a dataset is marked in bold.}
    \label{tab:results_real_world}
\end{table*}

\paragraph{Synthetic Datasets}

\begin{figure}[htbp]
    \centering
    \subfigure{
        \includegraphics[width=0.42\textwidth]{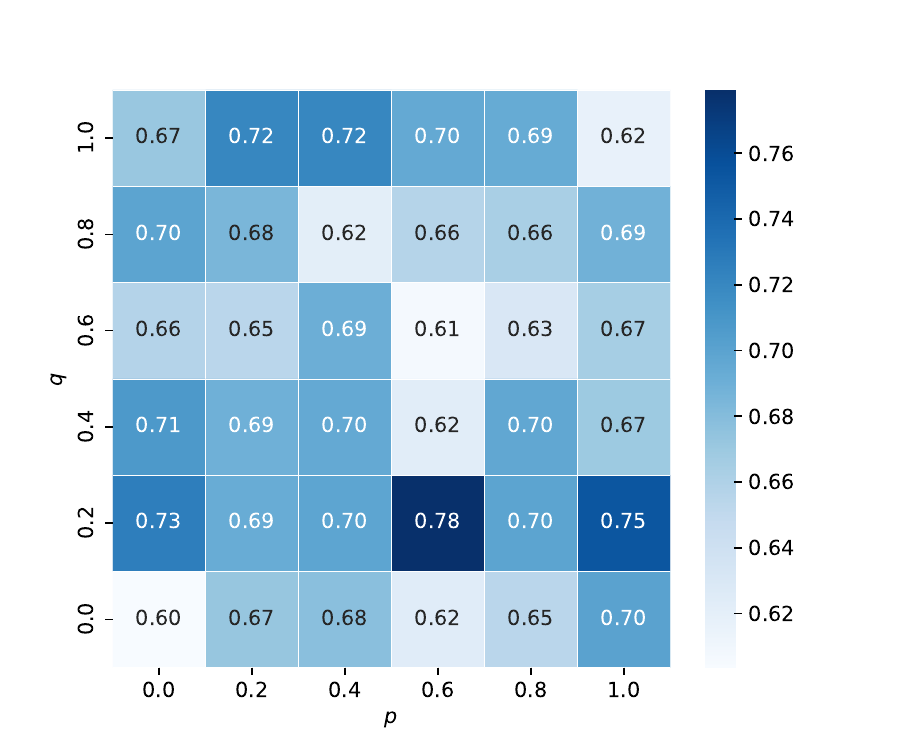}
        
        \label{fig:csbm heatmap}
    }
    \hspace{0.05\textwidth}
    \subfigure{
        \includegraphics[width=0.42\textwidth]{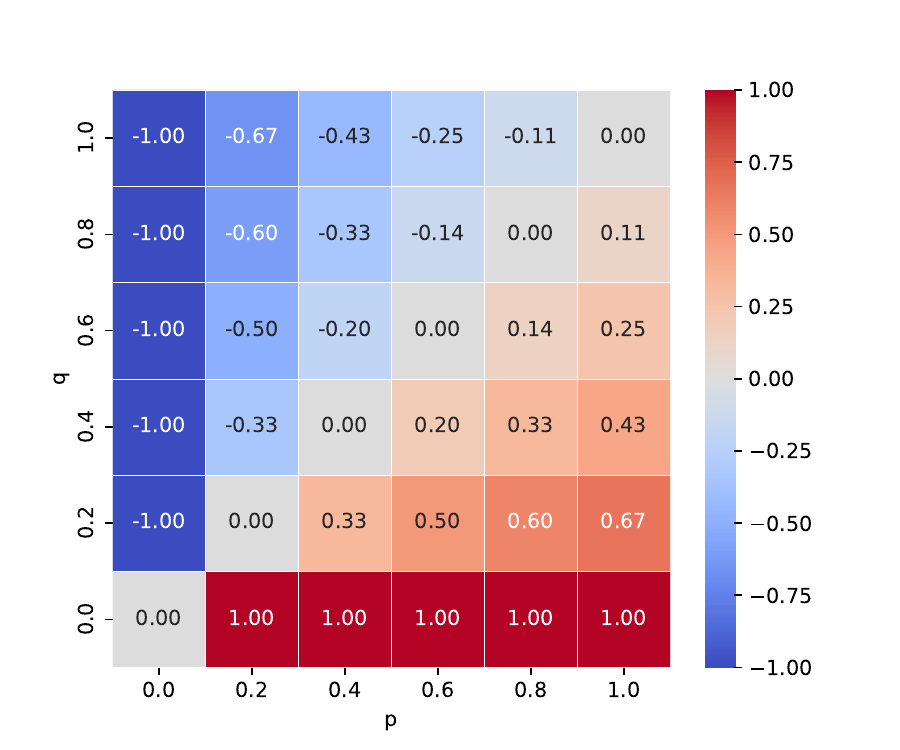}
        \label{fig:adjusted_homophily}
    }
    \caption{Test accuracies for the CSBM datasets (left). The scores are averaged over 5 runs.
            The probability for intra-class edges is on the x-axis and the probability for inter-class edges is on the y-axis. 
            The adjusted homophily ratios $\mathcal{H}_{adj}$ for the CSBM datasets (right).}
    \label{fig:CSBM_main_Heatmaps}
\end{figure}

Our results for the CSBM datasets are provided in Figure \ref{fig:csbm heatmap}.
The scores are averaged over $5$ runs.
We provide the adjusted homophily in Figure~\ref{fig:adjusted_homophily}.
ES-MLP performed best on the dataset with $p=0.6$ and $q=0.2$ with $78\%$.
The lowest score is $60\%$ for $p=0.0$ and $q=0.0$ \ie with no edges. 
For $p=1.0$ and $q=1.0$ we get a similar score of $0.62$.

\paragraph{Inference Time}
The results for the inference times are presented in Table~\ref{tab:InferenceTimes}.
We measure the times in two settings.
Either the model is evaluated on the full graph (left) or the model is evaluated on the test nodes only (right).
This is an important difference since the MLP-based models only require the test nodes and do not need the edge information, unlike the MPNN models.
The scores are given in milliseconds and averaged over $10$ runs.
All MLP-based models achieve lower inference times than the MPNNs. 
ES-MLP is two to five times faster in inference than the fastest MPNN.
From MPNNs, GraphSAGE was at least two times faster than ES-GNN, which is the slowest.
MPNNs lose up to $10\%$ classification accuracy when given test nodes only, due to the missing edges to non-test nodes.
MLP-based models are unaffected by this problem since they do not need the edges during inference.
The detailed test accuracies on test nodes only are in Appendix~\ref{appendix: extendet accuracies}.

\begin{table*}[!ht]
\adjustbox{max width=\textwidth}{
    \begin{tabular}{lrrrrrr|r|rrrrrr|r}
    \toprule
    & \multicolumn{6}{c|}{Inference time on test nodes using the full graph} & \multicolumn{6}{c}{Inference time on test nodes using only the test graph} \\
    \midrule
    & MLP & GCN & Graph- & ES- & Graph- & LINKX & ES- & MLP & GCN & Graph- & ES- & Graph- &LINKX& ES- \\ 
    &     &     & MLP    & GNN & SAGE   &       & MLP &     &     & MLP    & GNN & SAGE   &     & MLP \\ 
    \midrule
    Cora           & 0.150  & 0.963 & 0.206 & 1.389 &0.613 & 1.089 & 0.286 & 0.130 & 0.916 & 0.172 & 1.334  & 0.557 & 0.881 & 0.254 \\ 
    PubMed         & 0.150  & 1.763 & 0.219 & 1.367 &1.313 & 1.145 & 0.306 &  0.124 & 1.282 & 0.219 & 1.313 & 0.561 & 0.900 & 0.271 \\ 
    CiteSeer       & 0.144  & 1.230 & 0.222 & 1.381 &1.381 & 1.011 & 0.321 & 0.117 & 0.917 & 0.222 & 1.299 & 0.553 & 0.923 &0.264 \\ 
    \midrule
    Actor          & 0.166  & 0.996 & 0.243 & 1.395 &1.333 & 1.169 & 0.326 & 0.133 & 0.943 & 0.184 & 1.333 & 0.570 & 0.913 &0.290 \\ 
    Amazon          & 0.149  & 1.149 & 0.211 & 1.400 &1.322 & 1.194 & 0.314 & 0.128 & 0.918 & 0.179 & 1.322 & 0.556 & 0.918 & 0.273 \\ 
    Roman           & 0.141  & 1.554 & 0.204 & 1.392 &1.333 & 1.209 &0.301  & 0.127 & 0.897 & 0.174 & 1.333 & 0.556 & 0.915 & 0.261 \\ 
    Minesweeper    & 0.139  & 1.037 & 0.292 & 1.392 &1.289 & 1.253 & 0.294 & 0.109 & 0.913 & 0.292 & 1.289 & 0.624 & 0.911 &0.250 \\ 
    \bottomrule
\end{tabular}}
    \caption{The inference times in milliseconds on the full graph (left) and test graph only (right). The times are averaged over 10 runs.}
    \label{tab:InferenceTimes}
\end{table*}

\begin{figure*}[!ht]
    \includegraphics[width=\textwidth]{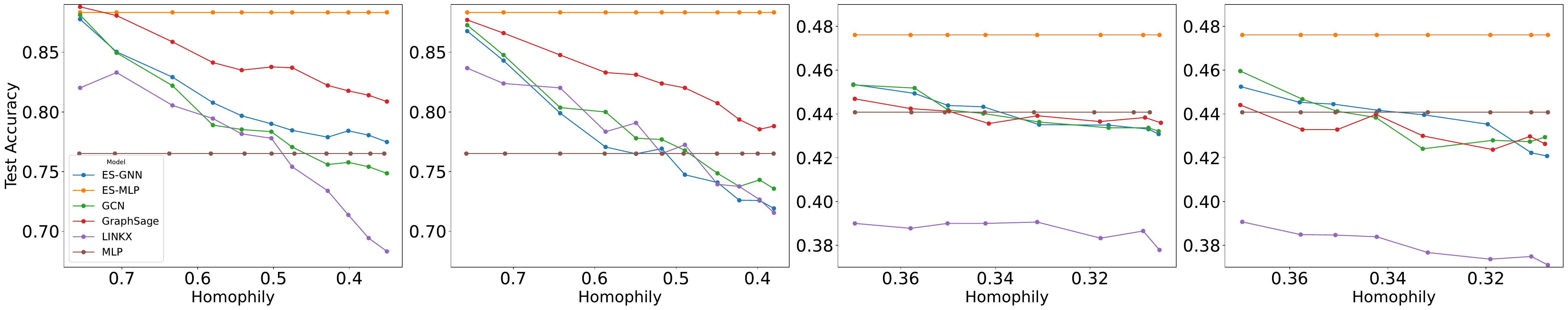}
        \caption{Test accuracies of the robustness analysis on the Cora and Amazon datasets. 
        Left to right: Results on Cora with uniform edge-noise, Cora with categorical edge-noise, Amazon with Uniform edge-noise, and Amazon with categorical edge-noise. 
        Our ES-MLP, as it does not require edges at test time, remains constant on all datasets.
        Graph-MLP, not shown for brevity, also remains constant. 
        }
    \label{fig:Robustness Analysis}
\end{figure*}

\paragraph{Robustness Analysis}
Results for the robustness analysis are shown in Figure~\ref{fig:Robustness Analysis}.
For the robustness analysis, we added edge noise during test time to observe how models perform when neighborhood distribution changes on the test graph.
Our results show that MPNNs lose performance if neighborhood distribution on the test graph differs from the training graph.
In the uniform edge-noise setting, GraphSAGE's performance decreases up to $8\%$, GCN up to $12.7\%$, and ES-GNN up to $10.3\%$ on Cora.
For the heterophilic graph Amazon, all MPNNs lose between $2$ and $3\%$, while the MLP-based models stay stable.
We observe for both datasets that the decrease is larger when adding noise with a categorical distribution. 
The MLP-based models maintain performance for both edge-noise distributions, independent of the magnitude of the edge noise added to the test graph.

\section{Discussion}
\label{sec:discussion}

The results of our experiments show that ES-MLP is effective for both homophilic and heterophilic graphs.
Extending Graph-MLP by a second channel to split the edges based on relevant versus irrelevant features enables the model to improve its performance on heterophilic graphs.
Notably, this improvement comes without losing performance on homophilic graphs.
\add{An analysis of the learned adjacency matrices shows that the edge-splitting separates homophilic from heterophilic edges, see Appendix~\ref{appendix:adj_vis} for details.}

\paragraph{Real-world Datasets}
On homophilic datasets, ES-MLP achieves competitive results to all baselines. 
The difference to the best-performing models GCN and Graph-MLP is between $0.31$ and $2.07$~points.
We observe that ES-MLP effectively captures the relationship between node features and homophilic edges, maintaining performance on par with Graph-MLP. %
On heterophilic graphs, ES-MLP outperforms all baselines, except for Minesweeper.
ES-MLP shows strong improvements over Graph-MLP, for instance by up to $11.18$ points on Amazon, \ie edge-splitting effectively improves the performance for heterophilic models.
Graph-MLP cannot compete on the heterophilic dataset, since the original neighborhood contrastive loss is based on the homophily assumption, which is violated here.
This highlights the capacity of ES-MLP to generalize on heterophilic graphs.
All MLP-based models achieve on Minesweeper an AUROC score of about $50$ points.
This is due to the fact that the node features are independent of the class of a node.
The class of a node is solely determined by the features of the neighbors. 
Since MLP-based models do not have access to these features during inference time, they are unable to classify these nodes correctly.
\add{
Recent studies demonstrated that node classification performance degrades when
the local homophily of a node deviates from the global homophily~\cite{PerformanceDiscrepanciesLocalHomphily}.
Our findings show similar behavior for ES-MLP, see Appendix~\ref{appendix:local_vs_global_homophily} for details.}

\paragraph{Synthetic Graphs}
The experiments on the CSBM datasets show that the worst results are obtained for a graph with no edges ($p=0$, $q=0$).
The second worst results are obtained using a fully connected graph ($p=1$, $q=1$) since no information can be learned from the edges.
The results on the diagonal are all below $0.7$, since here the ratio of heterophilic and homophilic edges is equal, \ie there is only a low amount of information in the graph structure.
Compared to homophily ratios in Figure~\ref{fig:adjusted_homophily}, we see no strong relationship between the test accuracies and the homophily levels of the graphs.
Nevertheless, ES-MLP achieves stronger results when $p$ and $q$ tend to be more asymmetric.

\paragraph{Robustness Analysis}

Our results show that edge noise during test time leads to a performance decrease of MPNNs, while MLP-based models are unaffected.
The reason is that MLP-based models do not use edges during inference.
Adding uniform edge-noise to Cora leads to a performance decrease of up to $12.7\%$ and up to $20\%$  with categorical edge-noise.
Categorical noise is more challenging since the neighborhood distribution is shifted towards two specific classes, while for the uniform noise, the added neighbors are more likely to balance each other out. 
The effect is less for the Amazon dataset, as it is already quite heterophilic, \ie the relative change due to the noise is smaller than for the homophilic Cora dataset.

\paragraph{Ablation Study}
We perform an ablation study on the neighborhood contrastive loss $\mathcal{L_{NC}}$ and the irrelevant consistency regularization loss $\mathcal{L_{ICR}}$ to analyze the effect on the performance of ES-MLP.
The results of the ablation study are provided in Table \ref{table:ablation study} by mean test accuracy.
On every dataset, the scores decrease when dropping the neighborhood contrastive loss.
The difference is highest for Roman with $4.52\%$.
This implies that the loss is important for training ES-MLP.
The homophilic datasets CiteSeer and PubMed achieved their best scores with $\beta_{ICR}$ of $0$, \ie the accuracies remain the same without the ICR loss.
Therefore, the ICR loss does not have a beneficial impact on the homophilic datasets, which was also the result of our hyperparameter search. 
On heterophilic datasets, the performance decreases when the ICR loss is dropped. 
ES-MLP loses $1.89$, $3.92$, and $3.73$ points on Amazon, Roman, and Actor, \ie the ICR loss is important for heterophilic datasets.
A sensitivity analysis for both loss parameters $\alpha_{NC}$ and $\beta_{ICR}$ can be found in Appendix~\ref{appendix:hyper_sensitivity}.

\begin{table}[!th]
    \small
    \centering
    \begin{tabular}{l | r | r | r | r | r | r}
        \toprule

          & \textbf{Cora} &\textbf{CiteSeer} & \textbf{PubMed} & \textbf{Actor} & \textbf{Amazon} & \textbf{Roman} \\
        \hline  
        ES-MLP w/o $\mathcal{L_{NC}}$& $86.16_{0.95}$ & $72.84_{0.74}$ &  $86.62_{0.22}$ & $36.00_{0.09}$ & $45.22_{0.02}$ & $60.92_{0.15}$ \\
        ES-MLP w/o $\mathcal{L_{ICR}}$ & $87.15_{0.52}$ & $\mathbf{75.67_{0.63}}$ & $\mathbf{87.56_{0.14}}$ & $35.12_{0.33}$& $46.36_{0.14}$ & $61.52_{0.15}$ \\
        \hline
        ES-MLP &$ \mathbf{88.15_{1.85}}$ & $\mathbf{75.67_{0.92}}$ & $\mathbf{87.56_{1.23}} $ & $\mathbf{39.73_{0.37}}$ & $\mathbf{47.85_{1.23}}$ & $\mathbf{65.44_{0.92}}$  \\
        \bottomrule

    \end{tabular}
    \caption{Ablation study on separate parts of the loss functions. Scores are given as test accuracies. The best results are marked in bold.}
    \label{table:ablation study}
\end{table}

\paragraph{Limitations}
\label{sec:threattovalidity}
Although ES-MLP can learn a meaningful relationship between neighbors and features, it cannot be applied to tasks where the class depends on the neighborhood structure during inference time, as shown by the Minesweeper dataset.
In contrast to MPNNs, MLP-based methods like ES-MLP are unaware of neighborhood distributions~\cite{IsHomophilyANecessity}.

\section{Conclusion}
\label{sec:conclusion}
We proposed ES-MLP, a GNN for homophilic and heterophilic graphs. Our model combines the edge-splitting mechanism of ES-GNN with the neighborhood contrastive loss of Graph-MLP.
We found extending the neighborhood contrastive loss improves performance on heterophilic graphs, while not losing performance on homophilic graphs.
For future work, ES-MLP may be extended to model graph directionality like~\cite{EdgeDirectionalyForHeteroGraphs}, to improve its performance on heterophilic datasets.

\textit{Acknowledgement.} The paper is the result of the first author's BSc thesis and Master's project.
The authors acknowledge support from the state of Baden-Württemberg through bwHPC.

\newpage

\bibliographystyle{unsrtnat}
\bibliography{reference}

\newpage

\appendix
\section*{Appendix}
\label{appendix:supplementarymaterials}

\section{Hyperparameter Optimization}
\label{hyperparameter_optim}
The hyperparameter optimization consists of two steps, we first optimize the parameters for MLP, GCN, and GraphSAGE.
For the MLP-based models, Graph-MLP, and ES-MLP, we reuse the hyperparameters of MLP and tune the model-specific parameters.
For MPNN based model ES-GNN, we use the parameters of GCN and only tune the model-specific parameters that GCN does not have.

\subsection{MLP, GCN, and GraphSAGE}
We adopt the hyperparameter tuning from~\citet{IsHomophilyANecessity} and tune the learning rate, weight decay, dropout rate, and hidden dimension.
We conducted hyperparameter tuning for the baseline models GCN, MLP, GraphSAGE, and \add{LINKX}. 
The hyperparameter search space included the following ranges.
The learning rate was varied over $\{0.002, 0.005, 0.01, 0.05\}$, weight decay values were selected from $\{5e-04, 5e-05, 5e-06, 5e-07, 5e-08, 1e-05, 0\}$, dropout rates from $\{0, 0.2, 0.5, 0.8\}$, and hidden dimensions from $\{64, 128, 256\}$. 
The best hyperparameters are reported in Table~\ref{tab:HPMLP},~\ref{tab:HPGCN} and~\ref{tab:HPGS} for MLP, GCN, and GraphSAGE respectively.

\begin{table}[!ht]
    \small
    \centering
    \begin{tabular}{l r r r r}
        \toprule
         \textbf{Datasets}& Hidden & Learning rate & Dropout & Weight Decay \\
         \hline
         Cora & 256&0.05&0.8&$5\text{e-}04$\\
         CiteSeer & 64&0.05&0.8&$5\text{e-}07$\\
         PubMed & 256&0.05&0.5&$1\text{e-}05$\\
         \hline
         Chameleon & 64&0.01&0.8&$5\text{e-}07$\\
         Squirrel & 256&0.05&0.8&$5\text{e-}04$\\
         Actor &256&0.05&0.8&$5\text{e-}05$\\
         Amazon &128&0.05&0.2&$5\text{e-}07$\\
         Roman &128&0.05&0.8&$5\text{e-}05$\\
         Minesweeper &256&0.01&0.5&$5\text{e-}04$\\

         \bottomrule
    \end{tabular}
    \caption{Used Hyperparameters for MLP}
    \label{tab:HPMLP}
\end{table}

\begin{table}[!ht]
    \small
    \centering
    \begin{tabular}{l r r r r}
    \toprule
         \textbf{Datasets}& Hidden & Learning rate & Dropout & Weight Decay \\
         \hline
         Cora & 128 & 0.05 & 0.8 & $5\text{e-}04$ \\
         CiteSeer& 32 & 0.005 & 0.2 & $5\text{e-}04$ \\
         PubMed & 256 & 0.05 &  0.5 & $5\text{e-}07$\\
         \hline
         Chameleon & 64 &0.05&0.5&  $5\text{e-}07$\\
         Squirrel & 256 &0.005&0.8&$1\text{e-}05$\\
         Actor & 64&0.05&0.8&$5\text{e-}04$\\
         Amazon &128&0.05&0.2&0.0\\
         Roman & 256&0.05&0.5&$5\text{e-}07$\\
         Minesweeper & 128&0.005&0.8&$1\text{e-}07$\\
         \bottomrule
    \end{tabular}
    \caption{Used hyperparameters for GCN}
    \label{tab:HPGCN}
\end{table}

\begin{table}[!ht]
    \small
    \centering
    \begin{tabular}{l r r r r}
    \toprule
         \textbf{Datasets}& Hidden & Learning rate & Dropout & Weight Decay \\
         \hline
         Cora&256&0.005&0.5&$5\text{e-}04$\\
         CiteSeer&256&0.05&0.8&$5\text{e-}04$\\
         PubMed&128&0.05&0.5&$1\text{e-}05$\\
         \hline
         Chameleon&64&0.01&0.8&0.0\\
         Squirrel&256&0.05&0.8&$5\text{e-}04$\\
         Actor&64&0.05&0.8&$5\text{e-}06$\\
         Amazon&256&0.01&0.2&0.0\\
         Roman&256&0.05&0.2&$5\text{e-}06$\\
         Minesweeper&128&0.005&0.8&$5\text{e-}07$\\ 
         \bottomrule
    \end{tabular}
    \caption{Used hyperparameters for GraphSAGE}
    \label{tab:HPGS}
\end{table}

\begin{table}[!ht]
    \small
    \centering
    \begin{tabular}{l r r r r}
    \toprule
         \textbf{Datasets}& Hidden & Learning rate & Dropout & Weight Decay \\
         \hline
         Cora          &256&0.05&0.2&$5\text{e-}04$\\
         CiteSeer      &128&0.5&0.0&$5\text{e-}04$\\
         PubMed        &64&0.01&0.2&$5\text{e-}04$\\
         \hline
         Chameleon     &256&0.002&0.5&$5\text{e-}04$\\
         Squirrel      &64&0.002&0.8&$5\text{e-}04$\\
         Actor         &256&0.05&0.2&$5\text{e-}04$\\
         Amazon        &128&0.01&0.5&$5\text{e-}05$\\
         Roman         &256&0.01&0.8&$1\text{e-}05$\\
         Minesweeper   &256&0.01&0.8&$5\text{e-}04$\\ 
         \bottomrule
    \end{tabular}
    \caption{Used hyperparameters for LINKX}
    \label{tab:HPLINX}
\end{table}

\subsection{ES-GNN, Graph-MLP, and ES-MLP}
We tune the model-specific hyperparameters.
For the model-specific parameters, we optimize the weighting coefficient $\alpha_{NC}$, irrelevant consistency coefficient $\beta_{ICR}$, scaling parameter $\epsilon_R$ and $\epsilon_{IR}$, and powers of the adjacency matrix $r$.
We determine the hyperparameter the same way as it is done by \citet{GraphMLP:abs-2106-04051} and \citet{Edge-SplittingGNN} by using Grid-Search.
We determine $\alpha_{NC}$  in a range of $\{0, 1, 10, 100\}$, $\beta_{ICR}$ in $\{1e-5, 1e-4, 1e-3, 1e-1\}$, $\epsilon_R$ and $\epsilon_{IR}$, in $\{0, 0.1, 0.3, 0.5, 0.7\}$, and the powers of the adjacency matrix $r$ in $\{1, 2, 3\}$.

For each dataset, we tuned the hyperparameters as described in Section~\ref{sec:hyperparameteroptimization}.
The Tables~\ref{appendix:hyperparameteroptimization_esmlp},~\ref{appendix:hyperparameteroptimization_graphmlp} and~\ref{appendix:hyperparameteroptimization_esgnn}below provide the hyperparameter settings for ES-MLP, Graph-MLP, and ES-GNN respectively, we used for each dataset.

\begin{table}[!ht]
    \small
    \centering
    \begin{tabular}{ l  r  r  r r  r }
    \toprule
         \textbf{Dataset} & adjacency power $r$ & \textbf{$\alpha_{NC}$}  & \textbf{$\beta_{ICR}$}  & \textbf{$\epsilon_R$} & \textbf{$\epsilon_{IR}$}\\
         \hline
         Cora       & $4$  & $1$ & $0.01$ & $0.5$ & $0.5$ \\
         CiteSeer   & $2 $ & $1$  & $0.0$   & $0.1$ & $0.3$ \\
         PubMed     & $2$  &  $1$ &  $0.0$  &   $0.3$  & $0.1$ \\
         \hline
         Chameleon &  $2$ & $10$ &  $0.001$    &  $0.3$  &   $0.5$   \\
         Squirrel  &  $1$ & $1$ & $0.0001$ &  $0.7$ &    $0.3$   \\
         Actor &3&1&0.0001&0&0.7\\
         Amazon  &3&1&0.0001&0&0.5\\
         Roman &1&1&0.00001&0.5&0.3\\
         Minesweeper &4&100&0.01&0.7&0.7\\
         \bottomrule
    \end{tabular}
    \caption{Hyperparameters for ES-MLP}
    \label{appendix:hyperparameteroptimization_esmlp}
\end{table}

\begin{table}[!th]
    \small
    \centering
    \begin{tabular}{l  r  r }
        \toprule
         \textbf{Dataset} & adjacency power $r$ & \textbf{$\alpha_{NC}$} \\
         \hline
         Cora     & $2$ & $10$ \\
         CiteSeer & $2$ & $1$ \\
         PubMed   & $1$ & $10$ \\
         \hline
         Chameleon & $2$ & $10$ \\
         Squirrel & $1$ & $10$ \\
         Actor & - & $0$ \\
         Amazon &-& $0$\\
         Roman &1&1\\
         Minesweeper &2&10\\
         \bottomrule
    \end{tabular}
    \caption{Hyperparameters for GraphMLP}
    \label{appendix:hyperparameteroptimization_graphmlp}
\end{table}

\begin{table}[!ht]
    \small
    \centering
    \begin{tabular}{ l  r  r  r  }
    \toprule
         \textbf{Dataset} & \textbf{$\beta_{ICR}$}  & \textbf{$\epsilon_R$} & \textbf{$\epsilon_{IR}$}\\
         \hline
         Cora       & $0.0001$ & $0.7$ & $0.1$ \\
         CiteSeer   & $0.01$   & $0.7$ & $0.7$ \\
         PubMed     &$0.01$ &   $0.7$  & $0.3$ \\
         \hline
         Chameleon & $0.001$    &  $0.3$  &   $0.3$   \\
         Squirrel  & $0.01$ &  $0.7$ &    $0.3$   \\
         Actor & 0.0001 & 0.7 & 0.7\\
         Amazon  & 0.01 &0.7 & 0.5.7\\
         Roman & 0.00001 & 0.1 & 0.0\\
         Minesweeper & 0.0001 & 0.7 & 0.0\\
         \bottomrule
    \end{tabular}
    \caption{Hyperparameters for ES-GNN}
    \label{appendix:hyperparameteroptimization_esgnn}
\end{table}

\section{Homophily}
We considered four homophily measures. One based on edge level, one on node level, a class insensitive homophily, and an adjusted homophily ratio.
The first three measures indicate the homophily level in a range from 0 to 1, where 1 denotes a completely homophilic graph.

Edge Homophily Ratio, which is defined as follows:
\cite{BeyondHomophilyInGNN}
 \begin{equation}
    h = \frac{|{(u,v): (u,v) \in{E} \wedge y_u = y_v}|}{|E|}
\end{equation}

This ratio indicates the proportion of edges in a graph that connects nodes with the same class (intra-edges). 

Node Homophily Ratio \cite{NodeHomoRatio} is defined as:

\begin{equation}
    \beta = \frac{1}{|V|} \sum_{v \in V} \frac{|{(w,v) : w \in N(v) \wedge y_v = y_w}|}{|N(v)|}
\end{equation}

It measures the ratio of intra-edges of the respective nodes in its neighborhoods, normalized over the whole number of nodes.

Class insensitive edge homophily Ratio\cite{ClassInsEdgeHomoRatio}

\begin{equation}
    \hat{h} = \frac{1}{C-1} \sum_{k=0}^{C-1} \max \left(0, h_k - \frac{|C_k|}{|V|}\right).
\end{equation}

Where $|C_k|$ denotes the number of nodes with class $k$, and $h_k$ is the class-wise homophily metric, given with 
\begin{displaymath}
    h_k = \frac{\sum_{u \in C_k d_u^{k_u}}}{\sum_{u \in C_k d_u}},
\end{displaymath}

where $d_u$ is the number of neighbors of node $u$ and $d_u^{k_u}$ the number of neighbors of $u$ that have the same class label. 
In this measure edge homophily is modified to be insensitive to the number of classes and size of each class. $\hat{h}$ measures presence of homophily

The adjusted homophily ratio~\cite{AdjustedHomophily} is formulated as:

\begin{displaymath}
    h_{adj} = \frac{h_{edge} - \sum_{k=1}^C \bar{p}(k)^2}{ 1 - \sum_{k=1}^C \bar{p}(k)^2},
\end{displaymath}

where $\bar{p}(k) = \frac{D_k}{2|E|}$ is the degree-weighted distribution of class labels.

\label{appendix: homophily measures}

\section{Accuracies of Inference Time Experiment}
\label{appendix: extendet accuracies}

We report the accuracies of our inference time experiment. 
We measured the accuracy for two settings.
Either the model is evaluated on the full graph or the model is evaluated on the test nodes only.
Therefore, the accuracies of the setting evaluated on the full graph correspond to the accuracies reported in Table~\ref{tab:results_real_world}.
The respective inference times can be found in Table~\ref{tab:InferenceTimes}. 
We also added Chameleon and Squirrel datasets~\cite{NodeHomoRatio} as additional experiments. 
These datasets are not included in the main paper, since they contain duplicate nodes as shown by \citet{NewHeteroData}.
The accuracies of the inference time experiment are reporter in Table~\ref{tab:Extendet Results}.

We observe, that by just providing the model with the test graph only, MPNN performance decreases since adjacent vertices from training are not present anymore.
MLP-based models are unaffected since they do not use edges during inference.
Therefore, MLP-based models can only be provided with the test graph to improve needed inference time while maintaining accuracy.
\add{This does not apply to LINKX. 
Although LINKX is an MLP-based model, it still requires edges during inference. 
For this reason, its performance degrades when applied only to the test graph.}

\begin{table*}[!ht]
\adjustbox{max width=\textwidth}{
    \begin{tabular}{lrrrrrr|r|rrrrrr|r}
    \toprule
    & \multicolumn{6}{c|}{Accuracies on test nodes using the full graph} & \multicolumn{6}{c}{Accuracies on test nodes using only the test graph} \\
    \midrule
    & MLP & GCN & Graph- & ES- & Graph- & LINKX & ES- & MLP & GCN & Graph- & ES- & Graph- & LINKX & ES- \\ 
    &     &     & MLP    & GNN & SAGE   &       & MLP &     &     & MLP    & GNN & SAGE   &       & MLP \\ 
    \midrule
    Cora         & $76.95_{1.00}$  & $\mathbf{88.46_{0.83}}$ & $86.64_{1.14}$ & $87.30_{0.43}$ & $88.26_{0.50}$ & $83.15_{0.59}$  & $88.15_{1.85}$ & $76.95_{1.00}$ &$80.03_{1.12}$ &  $86.64_{1.14}$ & $70.50_{6.71}$  & $79.49_{0.88}$ & $72.22_{1.19}$ &$\mathbf{88.15_{1.85}}$ \\ 
    CiteSeer       & $72.10_{1.12}$  & $77.41_{0.95}$          &$\mathbf{77.79_{0.10}}$ &  $74.27_{1.50} $ & $76.54_{0.73}$ &$73.23_{0.85}$& $75.67_{0.92}$ &   $72.10_{1.12}$ & $69.74_{0.64}$ & $\mathbf{77.79_{0.10}}$  & $62.24_{3.83}$  & $62.94_{3.74}$ & $70.93_{1.06}$ &$75.67_{0.92}$ \\ 
    PubMed     & $87.49_{0.90}$  & $\mathbf{89.63_{0.79}}$& $87.06_{2.41}$ & $88.81_{0.49}$ &$89.60_{0.41}$& $87.47_{0.29}$ &$87.56_{1.23}$ &$87.49_{0.90}$ & $85.52_{0.51}$ & $87.06_{2.41}$  &$82.56_{1.78}$ & $71.56_{1.78}$ & $87.54_{0.28}$ &$\mathbf{87.56_{1.23}}$ \\ 
    \midrule
    Chameleon  &$50.54_{1.05}$  & $46.36_{1.45}$            & $51.22_{0.14}$ & $66.32_{2.10}$ & $53.46_{0.99}$ & $\mathbf{68.10_{1.33}}$& $65.84_{2.19}$ & $50.54_{1.05}$ &$44.74_{1.23}$ & $51.22_{0.14}$ & $58.90_{1.25}$ & $26.90_{1.25}$ & $41.57_{1.65}$ &$\mathbf{65.84_{2.19}}$ \\ 
    Squirrel   & $34.76_{0.90}$  &$28.79_{1.00}$           & $34.18_{0.16}$ & $60.20_{0.92}$ &$35.54_{0.91}$ &$\mathbf{60.90_{0.81}} $& $55.01_{1.81}$ & $34.76_{0.90}$ & $28.62_{0.86}$ &  $34.18_{0.16}$  & $39.90_{0.94}$ & $21.90_{1.11}$ & $27.71_{1.56}$& $\mathbf{55.01_{1.81}}$ \\ 
    Actor       & $35.81_{0.62}$  &  $29.24_{0.47}$          & $36.03_{0.98}$ & $38.91_{0.45}$ &$32.24_{0.76}$ & $33.92_{1.11}$ &$\mathbf{39.73_{0.37}}$ &  $35.81_{0.62}$ &$30.76_{0.61}$ & $36.03_{0.98}$ & $31.29_{1.67}$ & $25.29_{1.68}$ &  $34.82_{1.22}$& $\mathbf{39.73_{0.37}}$\\ 
    Amazon      &$44.05_{0.54}$  &  $46.27_{0.67}$              & $37.07_{0.80}$ & $46.53_{0.34}$ &$44.83_{1.16}$ & $39.25_{0.51}$ &$\mathbf{47.8  5_{1.23}}$ &  $44.05_{0.54}$& $43.53_{0.78}$ &$37.07_{0.80}$  & $41.05_{0.68}$ & $41.05_{0.68}$ &  $39.26_{0.55}$& $\mathbf{47.8  5_{1.23}}$ \\ 
    Roman       & $60.50_{0.88}$  & $41.40_{1.58}$          & $64.94_{0.25}$& $60.41_{1.90}$ &$62.47_{1.90}$ & $65.40_{0.37}$ &$\mathbf{65.44_{0.92}}$ &$60.50_{0.88}$ & $42.10_{0.54}$ & $64.94_{0.25}$ & $52_.02_{0.09}$& $52.02_{0.09}$ &$64.61_{0.33}$& $\mathbf{65.44_{0.92}}$ \\ 
    Minesweeper    & $50.54_{0.49}$  &$71.44_{0.74}$            & $50.99_{0.35}$ & $68.23_{1.10}$ &$\mathbf{88.90_{2.37}}$ & $51.61_{1.4}$ &$ 50.87_{2.03}$ & $50.54_{0.49}$ & $\mathbf{62.01_{0.29}}$ &$50.99_{0.35}$ &  $56.85_{0.89}$ & $56.68_{0.89}$ & $48.96_{1.01}$& $ 50.87_{2.03}$ \\ 
    \bottomrule
\end{tabular}}
    \caption{Test accuracies and standard deviation in percent (\%) on the real-world datasets. Full denoting given the full graph as model input and sub denoting given the test graph only as model input. The best results are marked in bolt for both settings respectively.}
    \label{tab:Extendet Results}
\end{table*}

\section{Hardware}
\label{apppendix: hardware}
All experiments have been performed on a NVIDIA A100 GPU.
\label{ExtendetDSDescription}

\section{Additional Details for Section~\ref{sec:procedure}}
\label{appendix: distribution details}
We describe the details of how we added the edge-noise for the robustness analysis.
Specifically, we detail the distributions $D_{\bar{c}}$.

\paragraph{Cora}
Cora has seven classes, which we denote as $\{0, 1, 2, 3, 4, 5, 6\}$. 

\[
\mathcal{D}_0 : \text{Categorical}([0, 0.5, 0.5, 0, 0, 0, 0, 0]),
\]
\[
\mathcal{D}_1 : \text{Categorical}([0, 0, 0.5, 0.5, 0, 0, 0, 0]),
\]
\[
\mathcal{D}_2 : \text{Categorical}([0, 0, 0, 0, 0.5, 0.5, 0, 0]),
\]
\[
\mathcal{D}_3 : \text{Categorical}([0, 0, 0, 0, 0, 0.5, 0.5, 0]),
\]
\[
\mathcal{D}_4 : \text{Categorical}([0, 0, 0, 0, 0, 0, 0.5, 0.5]),
\]
\[
\mathcal{D}_5 : \text{Categorical}([0.5, 0, 0, 0, 0, 0, 0, 0.5]),
\]
\[
\mathcal{D}_6 : \text{Categorical}([0.5, 0.5, 0, 0, 0, 0, 0, 0]).
\]

\paragraph{Amazon}
Amazon has five classes, which we denote as $\{0, 1, 2, 3, 4\}$. 
\[
\mathcal{D}_0 : \text{Categorical}([0, 0.5, 0, 0, 0, 0, 0, 0.5]),
\]
\[
\mathcal{D}_1 : \text{Categorical}([0.5, 0, 0.5, 0, 0, 0, 0, 0]),
\]
\[
\mathcal{D}_2 : \text{Categorical}([0, 0.5, 0, 0.5, 0, 0, 0, 0]),
\]
\[
\mathcal{D}_3 : \text{Categorical}([0, 0, 0.5, 0, 0.5, 0, 0, 0]),
\]
\[
\mathcal{D}_4 : \text{Categorical}([0, 0, 0, 0.5, 0, 0.5, 0, 0]),
\]

\section{Extended Real-World Dataset Description}
In the following, we describe the datasets and their features in more detail since it may help to improve the understanding of some results.
For our experiments, we use nine datasets. Three datasets have a homophily
score above 0.5 and six datasets are heterophilic,i.e., homophily under 0.5. We measure homophily
by the adjusted homophily ratio~\cite{AdjustedHomophily}. This metric is suited to compare datasets with different numbers
of classes.
The homophilic datasets are the three standard citation graphs Cora, CiteSeer, and PubMed~\cite{CitationNetworkDatasets}. In
these datasets, each node represents a document and edges correspond to citations between them. In
Cora and CiteSeer, node features represent elements of a bag-of-words representation of a document
and the label of a node indicates the topic. In PubMed, features are represented by a TF-IDF weighted
vector. In these datasets, the nodes are more likely to be connected with nodes of the same class.
For the heterophilic datasets, where nodes are more likely to be connected with nodes from
other classes, we use Chameleon, Squirrel~\cite{NodeHomoRatio}, Actor~\cite{NodeHomoRatio}, Amazon, Roman, and
Minesweeper~\cite{NewHeteroData}. Chameleon and Squirrel are page-to-page graphs in Wikipedia with low ho-
mophily. Nodes represent Wikipedia articles and edges represent hyperlinks between pages. Node
features correspond to several informative nouns in the Wikipedia pages and the nodes are classified
into five categories, which denote the number of the average monthly traffic of the web page. In the
Actor dataset, the nodes correspond to an actor, and the edges between them denote co-occurrences
on the same Wikipedia page. Features correspond to some keywords in the Wikipedia pages. The
Roman dataset is based on the Roman Empire Wikipedia article. Each node corresponds
to one (non-unique) word in the text. Two nodes are connected with an edge, if either these words
follow each other in the text, or these words are connected in the dependency tree of the sentence. In
the Amazon dataset, nodes are products, and edges denote if products are frequently bought
together. Minesweeper dataset is inspired by the Minesweeper game. The graph is a $100 \times 100$ grid
where each node denotes a cell. Each cell is connected to 8 adjacent cells (except the cells at the
edge). Node features are one-hot-encoded numbers of adjacent mines. The task is to predict which
nodes are mines

\section{Edge Noise on Train and Test Set}

\begin{figure}[htbp]
    \centering
    \subfigure{
        \includegraphics[width=0.4\textwidth]{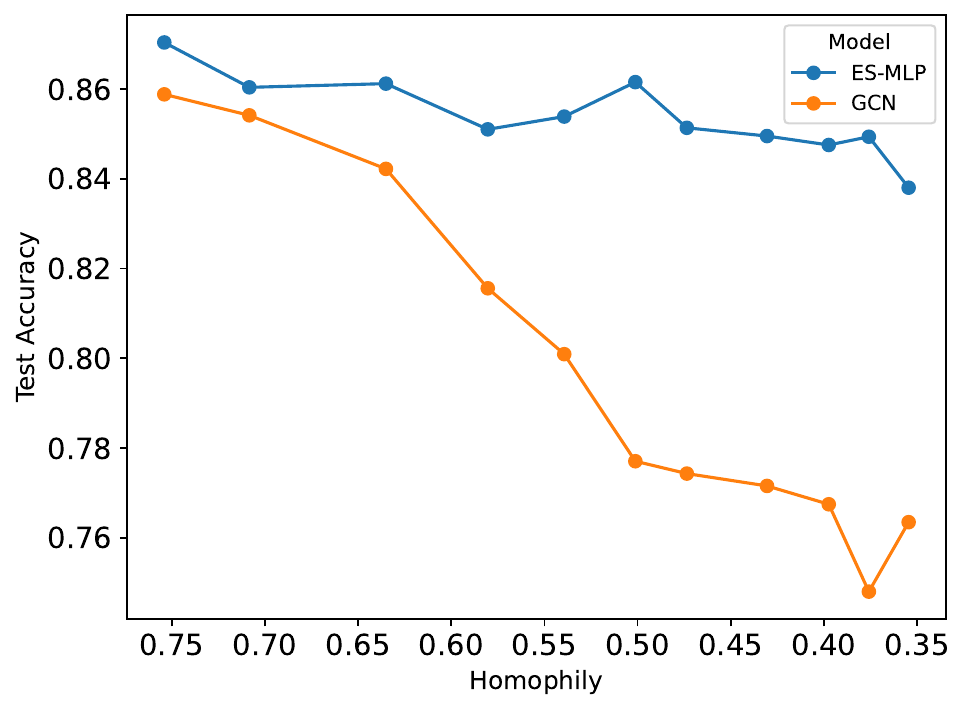}
        
        \label{fig:add_edge_noise_uniform}
    }
    \hspace{0.01\textwidth}
    \subfigure{
        \includegraphics[width=0.4\textwidth]{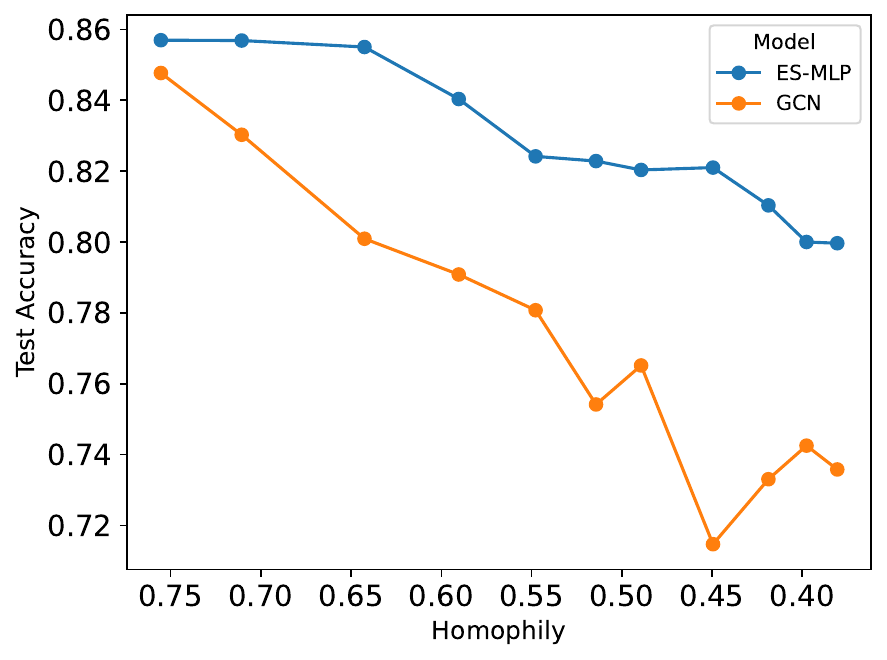}
        \label{fig:add_edge_noise_categorical}
    }
    \caption{Results for the edge-noise experiment on Cora with uniform (left) and categorical (right) noise.}
    \label{fig:add_edge_noise_exp}
\end{figure}

We performed the edge-noise experiment on the whole dataset, \ie the same noise during training and testing.
We added uniform noise and categorical noise to the existing edges on Cora. 
The test accuracy of the models with respect to the homophily level can be seen in Figure~\ref{fig:add_edge_noise_exp}. 
GCN loses up to $11.01\%$ accuracy for uniform edge-noise and up to $11.9\%$ with categorical noise, while ES-MLP only loses up to $3.24\%$ and $5.73\%$.
We observe that ES-MLP is more robust to edge-noise than GCN in both cases.
Note that in this setting, the edge-noise is already present during training, \ie the performance drop is not happening due to some distribution shift between train and test data.
Therefore, ES-MLP is not only robust towards edge noise during inference time but also towards general noisy edges in the data.

\section{Hyperparameter Sensitivity Analysis}
\label{appendix:hyper_sensitivity}
\begin{figure}[htbp]
    \centering
    \subfigure{
        \includegraphics[width=0.3\textwidth]{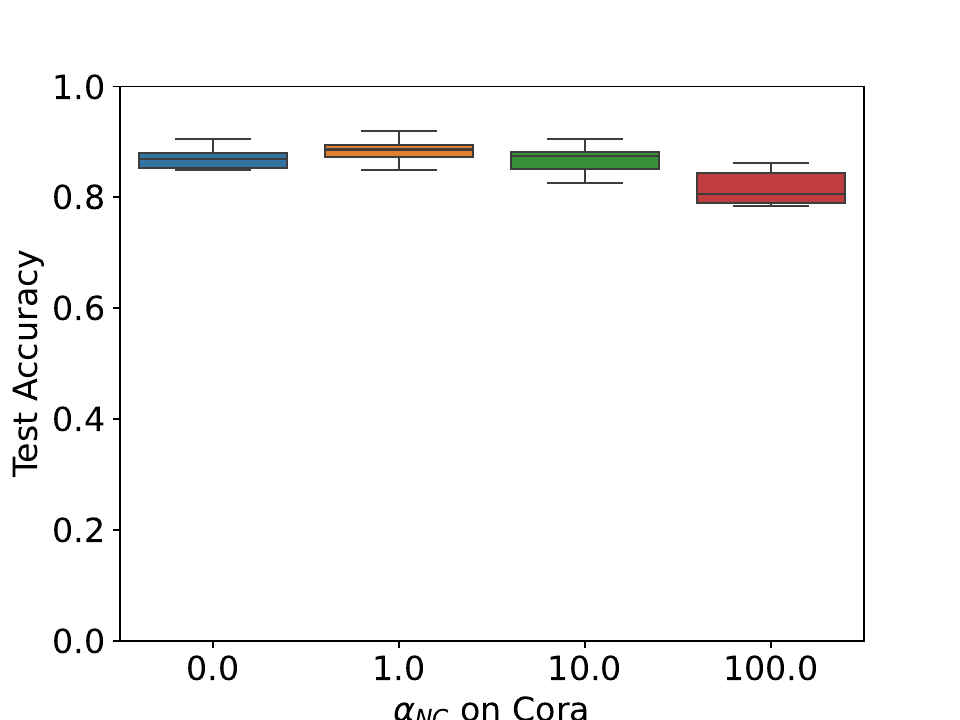}
        
        \label{fig:cora_alpha}
    }
    \hspace{0\textwidth}
    \subfigure{
        \includegraphics[width=0.3\textwidth]{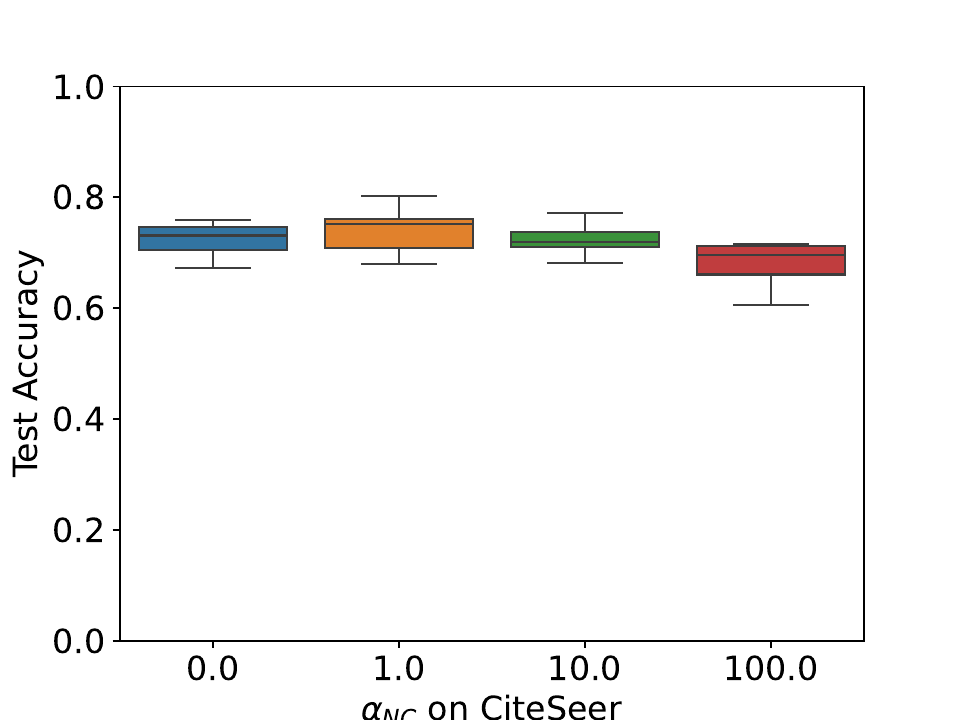}
        \label{fig:citeseer_alpha}
    }
    \hspace{0\textwidth}
    \subfigure{
        \includegraphics[width=0.3\textwidth]{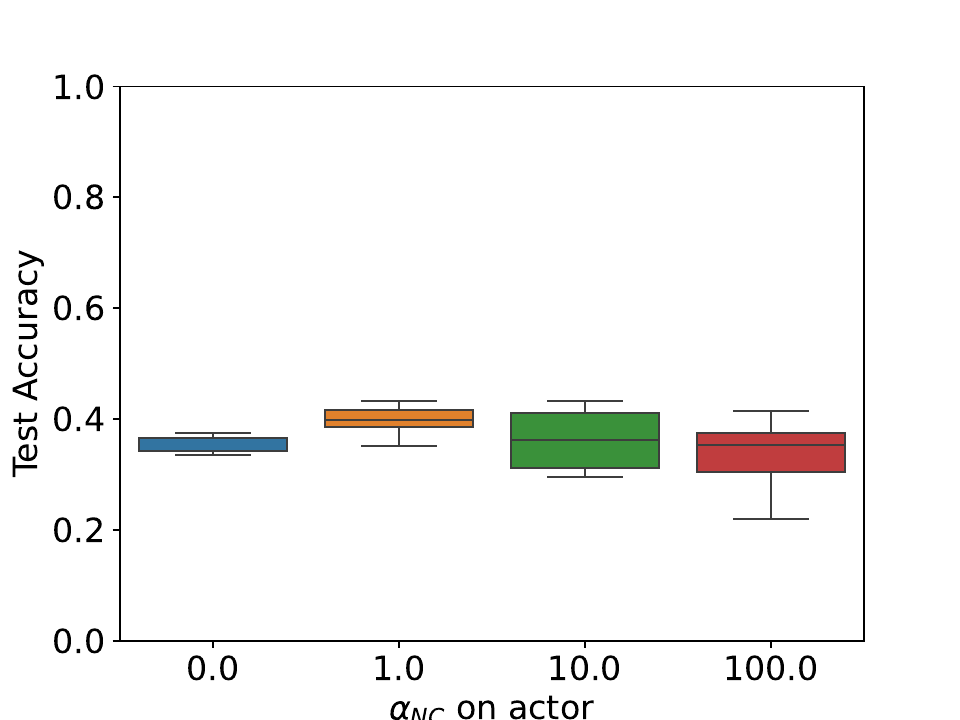}
        \label{fig:actor_alpha}
    }
    \hspace{0\textwidth}
    \subfigure{
        \includegraphics[width=0.3\textwidth]{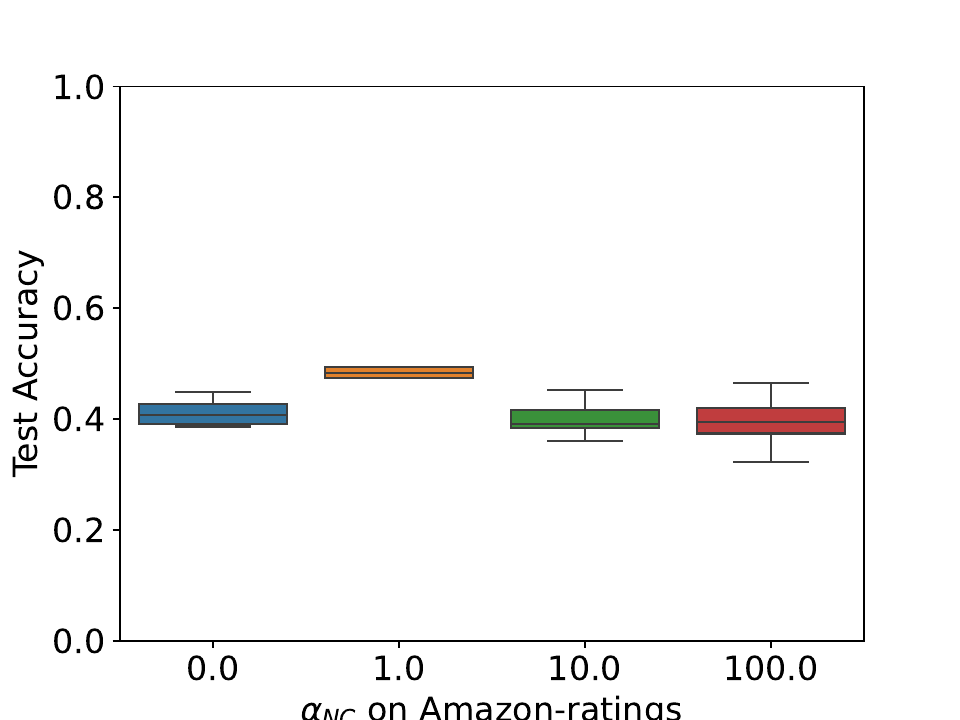}
        \label{fig:amazon_alpha}
    }
    \hspace{0\textwidth}
    \subfigure{
        \includegraphics[width=0.3\textwidth]{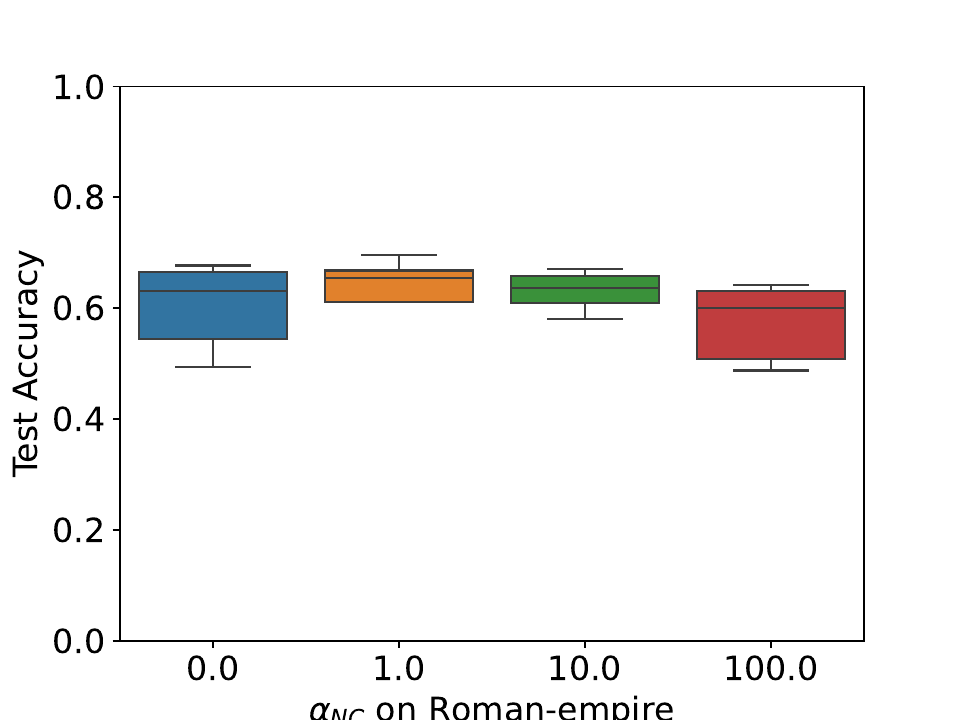}
        \label{fig:roman_alpha}
    }

    \caption{Hyperparameter sensitivity analysis on the neighborhood contrastive loss weight $\alpha_{NC}$ hyperparameter}
    \label{fig:LossWeightsAnalysisAlpha}
\end{figure}

\begin{figure}[htbp]
    \centering
    \subfigure{
        \includegraphics[width=0.3\textwidth]{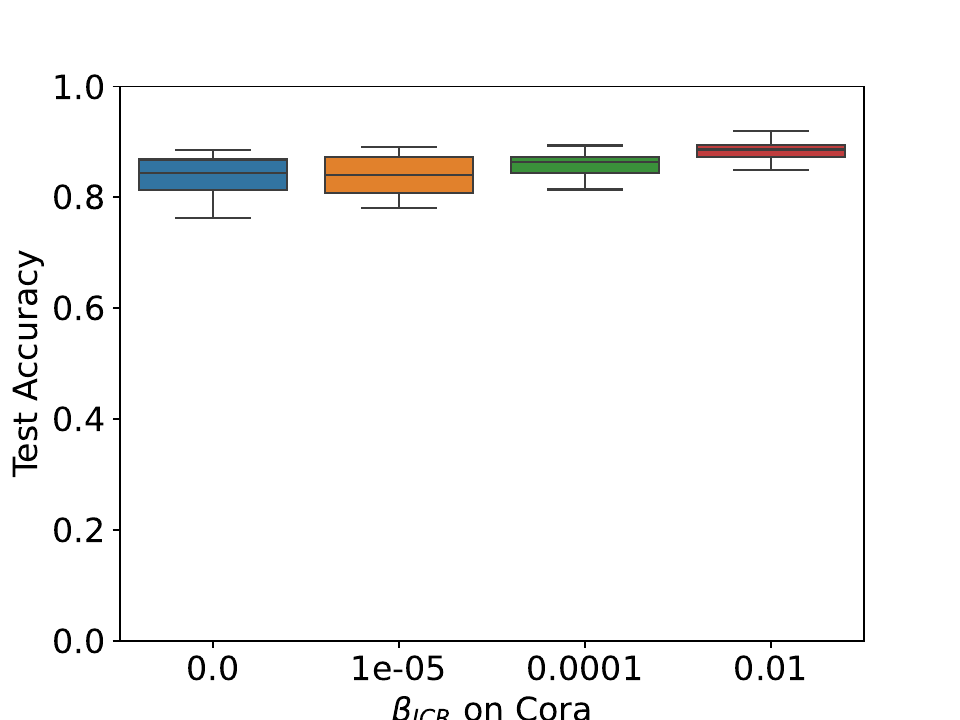}
        
        \label{fig:cora_beta}
    }
    \hspace{0\textwidth}
    \subfigure{
        \includegraphics[width=0.3\textwidth]{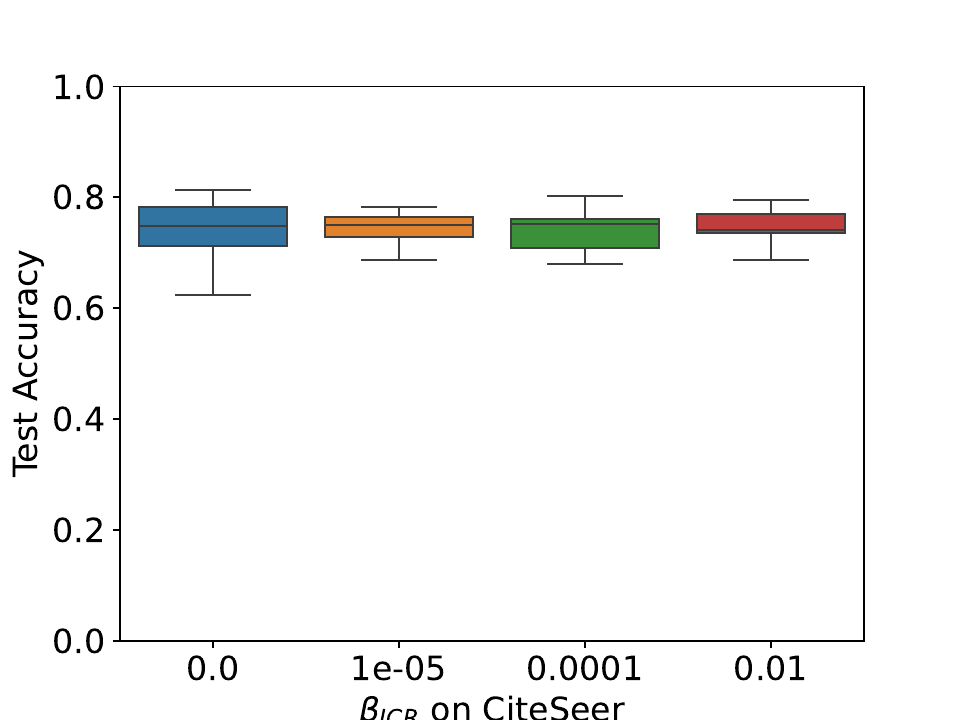}
        \label{fig:citeseeer_beta}
    }
    \hspace{0\textwidth}
    \subfigure{
        \includegraphics[width=0.3\textwidth]{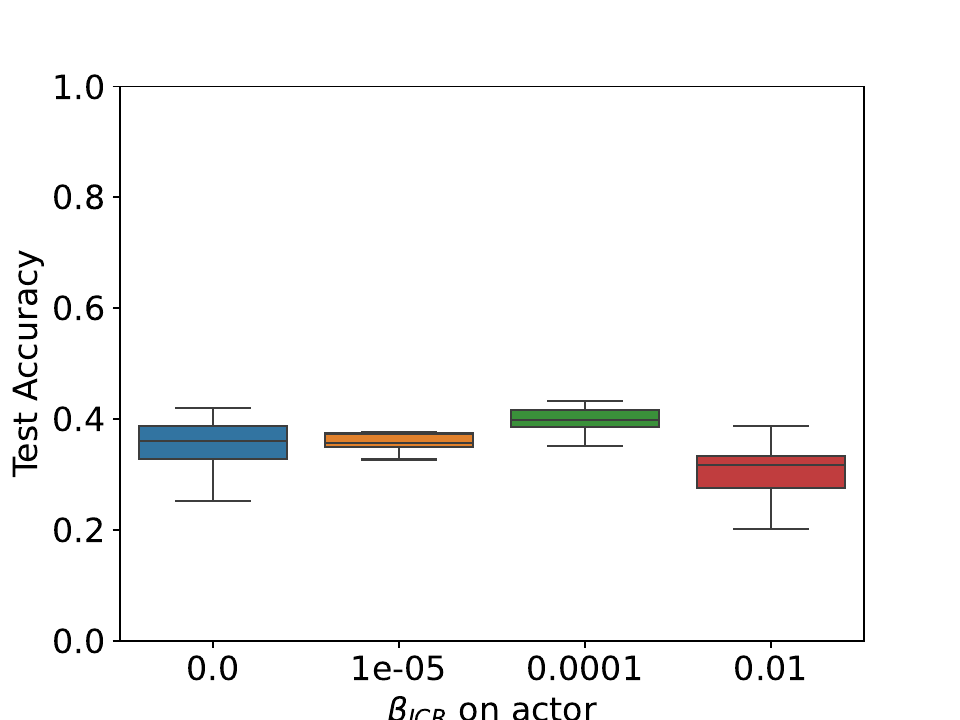}
        \label{fig:actor_beta}
    }
    \hspace{0\textwidth}
    \subfigure{
        \includegraphics[width=0.3\textwidth]{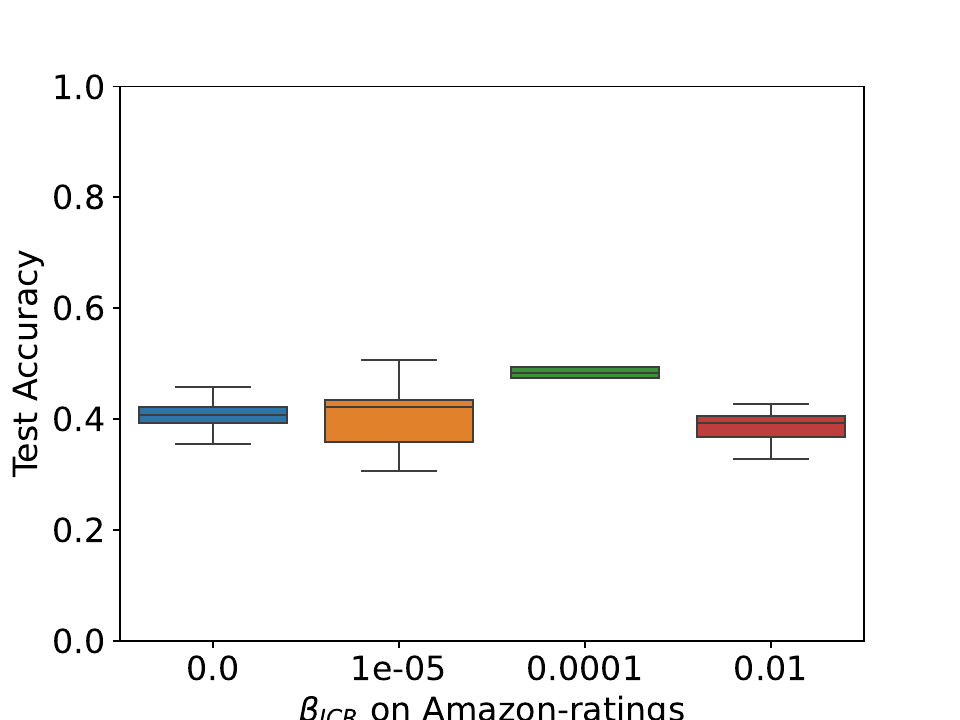}
        \label{fig:amazon_beta}
    }
    \hspace{0\textwidth}
    \subfigure{
        \includegraphics[width=0.3\textwidth]{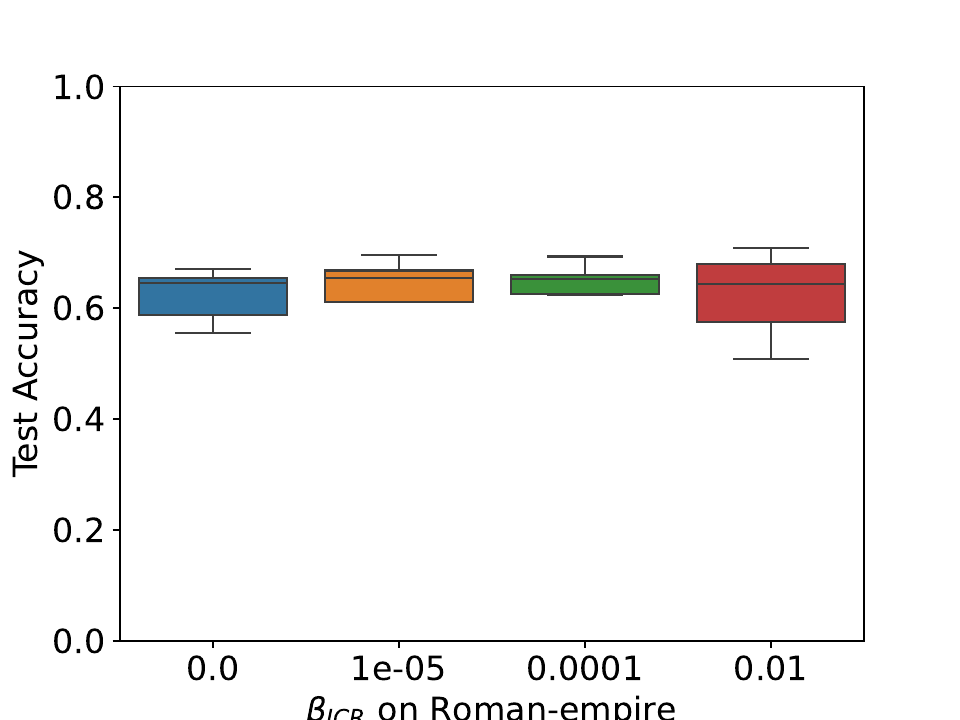}
        \label{fig:roman_beta}
    }

    \caption{Hyperparameter sensitivity analysis on the hyperparameter ICR loss weight $\beta_{ICR}$.}
    \label{fig:LossWeightsAnalysisBeta}
\end{figure}

\add{
We provide a hyperparameter sensitivity analysis for the neighborhood contrastive loss weight $\alpha_{NC}$ and the irrelevant consistency loss $\beta_{ICR}$ in Figures~\ref{fig:LossWeightsAnalysisAlpha} and~\ref{fig:LossWeightsAnalysisBeta}.
For the hyperparameter neighborhood contrastive loss weight, we observe that ES-MLP achieves the best performance when $\alpha_{NC}$ is $1$. On all datasets, the accuracy increases when the neighborhood contrastive loss $\alpha_{NC}$ is not $0$.
For higher values of $\alpha_{NC}$, the performance decreases again, since the model neglects the node features too much.
The irrelevant consistency loss weight $\beta_{ICR}$ is in general smaller than $\alpha_{NC}$.
Especially for heterophilic datasets, ES-MLP achieves the best performance when the ICR loss is not $0$.
Similar to $\alpha_{NC}$, the performance decreases on most datasets when $\beta_{ICR}$ is large.
Both parameters are quite robust, \ie by deviating the parameter values, the performance changes only slowly. 
This shows that most of the performance gain for heterophilic graphs is achieved by the edge-splitting mechanism in the architecture itself.}

\section{Local vs Global Homophily}
\label{appendix:local_vs_global_homophily}
\begin{figure}[ht!]
    \centering
    \includegraphics[width=1\linewidth]{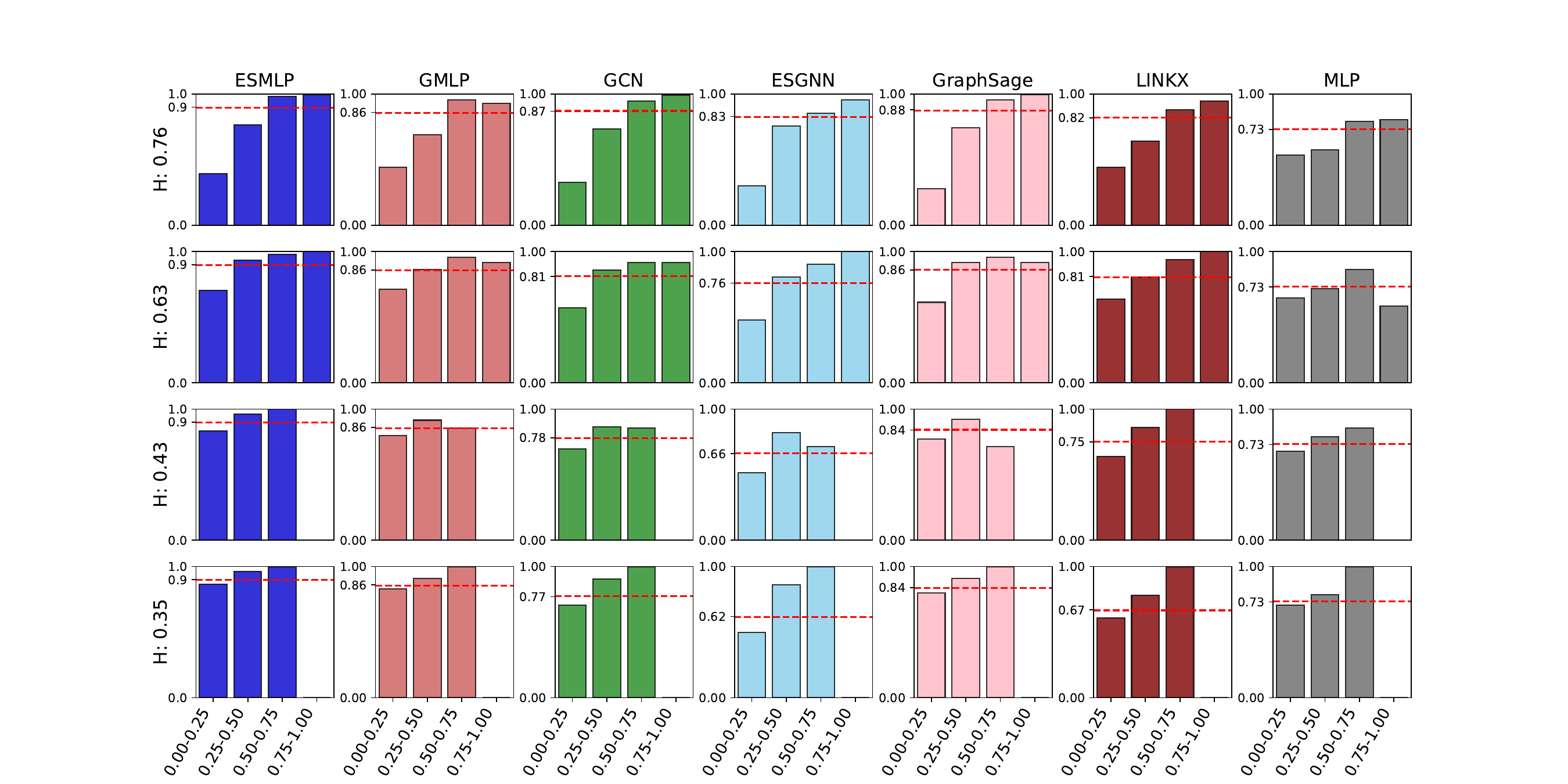}
    \caption{Performance for ES-MLP and baseline models on Cora with generated edge noise on the test nodes with global homophily ratios $\{0.35, 0.43, 0.58, 0.76\}$. Results are reported for different local homophily ranges.}
    \label{fig:local_vs_global_homophily}
\end{figure}

\add{
We analyze the performance discrepancies between local and global homophily ratios during the robustness analysis based on the work of \citet{PerformanceDiscrepanciesLocalHomphily}.
We trained the model once and evaluated it for multiple levels of edge noise.
We investigated the local homophily of single nodes for the global homophily level of $\{0.35, 0.43, 0.58, 0.76\}$ on Cora and $\{0.37, 0.36, 0.33, 0.31 \}$ on Amazon.
A bar plot for the accuracy of each quantile based on the nodes' homophily is shown in Figure~\ref{fig:local_vs_global_homophily}.
For Cora, we observe that the accuracy for heterophilic nodes increases as the global homophily reduces, while the homophilic nodes disappear in the second step which is consistent with the work of \citet{PerformanceDiscrepanciesLocalHomphily} and \citet{PAC-Bayesian}.
}

\section{Adjacency Matrix Visualization}
\label{appendix:adj_vis}
\begin{figure}[ht!]
    \centering
    \includegraphics[width=0.8\linewidth]{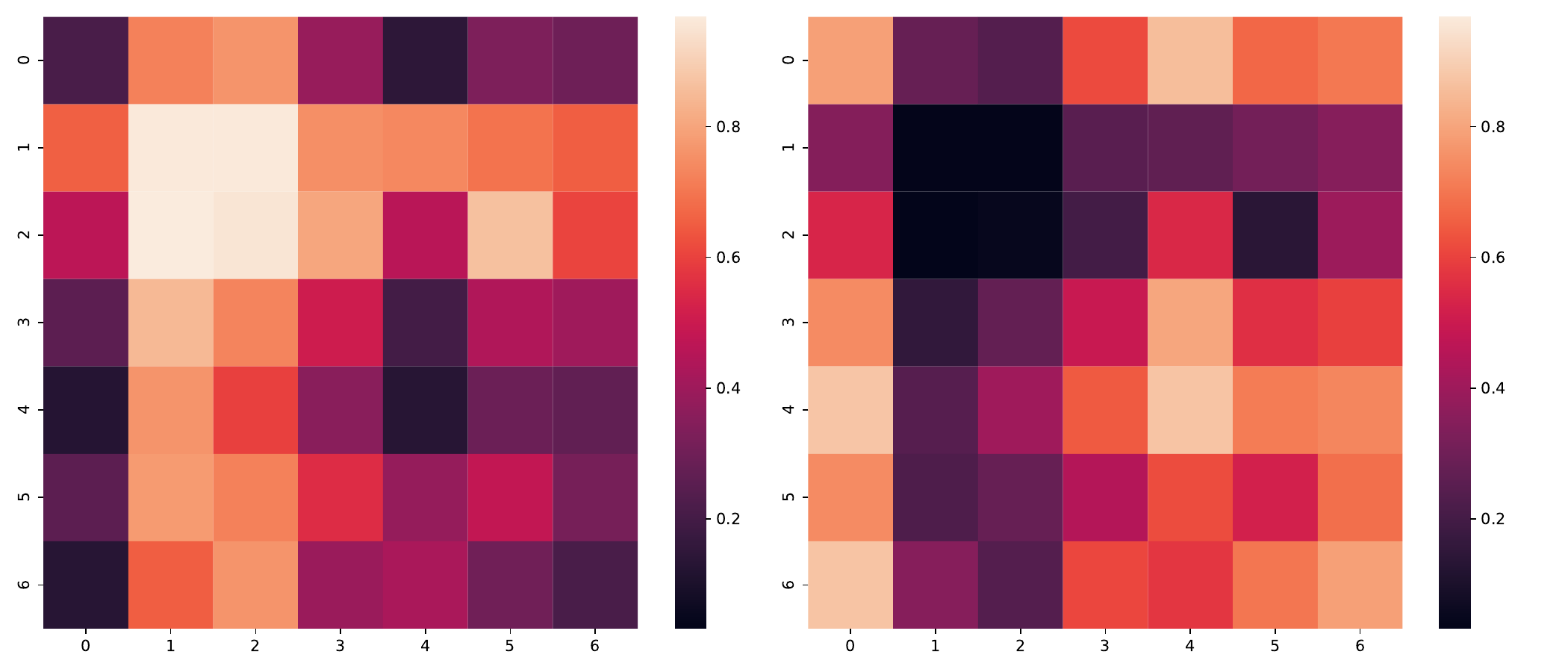}
    \caption{Adjacency Matrices $A_r$ and $A_{ir}$ on Cora. The ratio of homophilic edges and all edges in $A_r$ is $77.53\%$, and in $A_{ir}$ $66.82\%$ }
    \label{fig:edge_analysis_cora}
\end{figure}

\begin{figure}
    \centering
    \includegraphics[width=0.8\linewidth]{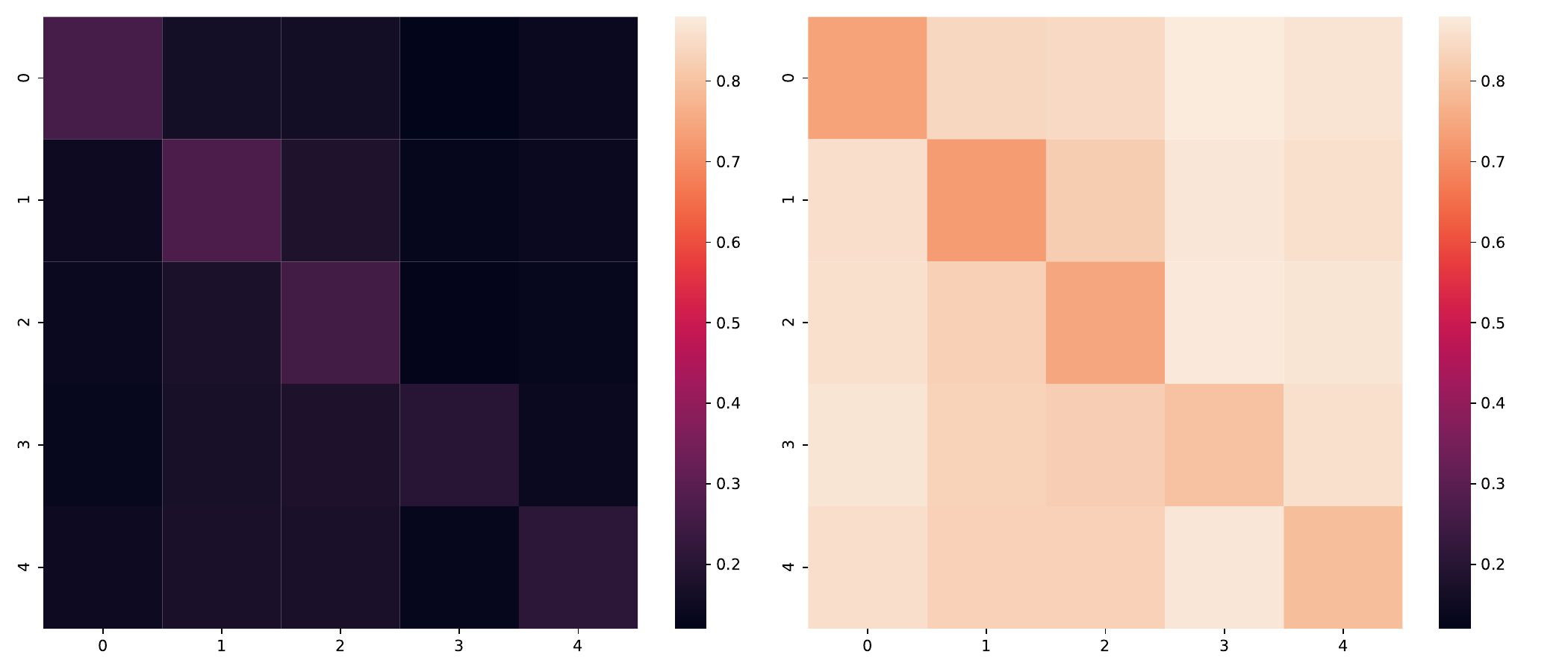}
    \caption{Adjacency Matrices $A_r$ and $A_{ir}$ on Actor. The ratio of homophilic edges and all edges in $A_r$  is $32.33\%5$, and in $A_{ir}$ $21.91\%$}
    \label{fig:edge_analysis_actor}
\end{figure}

\add{
We aggregated the $r$-th power of the $A_R$ and $A_{IR}$ matrix into a $|C|x|C|$ matrix $\mathcal{C}^s$, where $r$ is a hyperparameter used for the respective dataset. 
$\mathcal{C}_{ij}^s$ is computed by the ratio of edges between class $c_i$ and class $c_j$ in the $r$-th power of $A_R$ and $A_{IR}$ versus $A$.
The results can be seen in Figures~\ref{fig:edge_analysis_cora} and~\ref{fig:edge_analysis_actor}.}

\add{
We observe that $A_r$ has a higher ratio of edges on the diagonal than $A_{IR}$, \ie the learned adjacency matrix is more homophilic.
For Actor, the model learned to ignore most of the edges, \ie the majority of the edges is in $A_{IR}$.
}

\section{Extended CSBM Results}
\begin{figure}[htbp]
    \centering
    \subfigure{
        \includegraphics[width=0.4\textwidth]{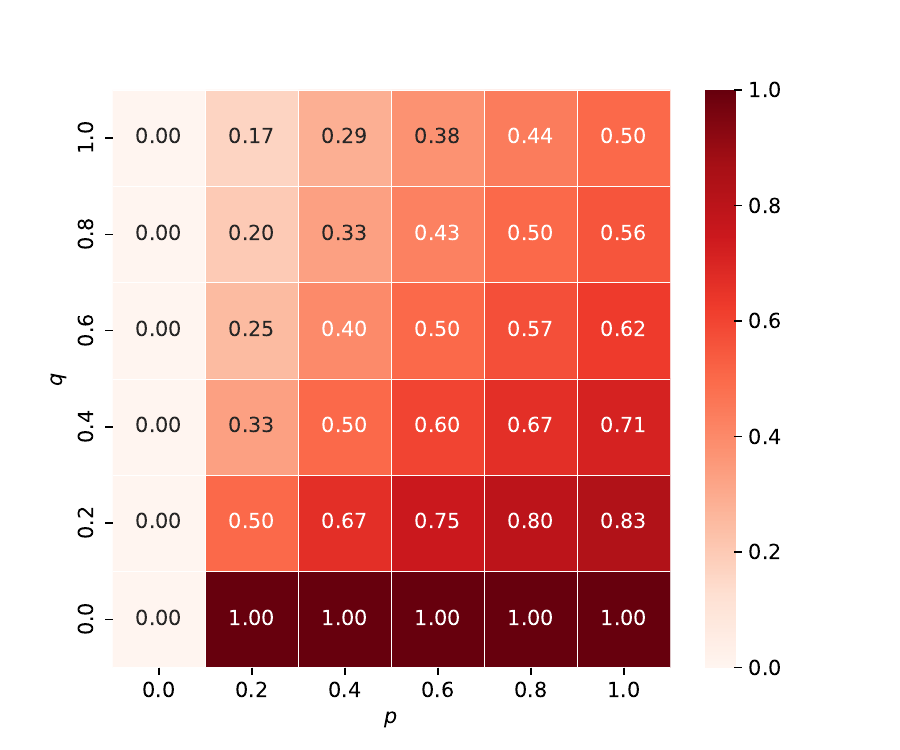}
        
        \label{fig:edge_homphily_heatmap}
    }
    \quad
    \subfigure{
        \includegraphics[width=0.4\textwidth]{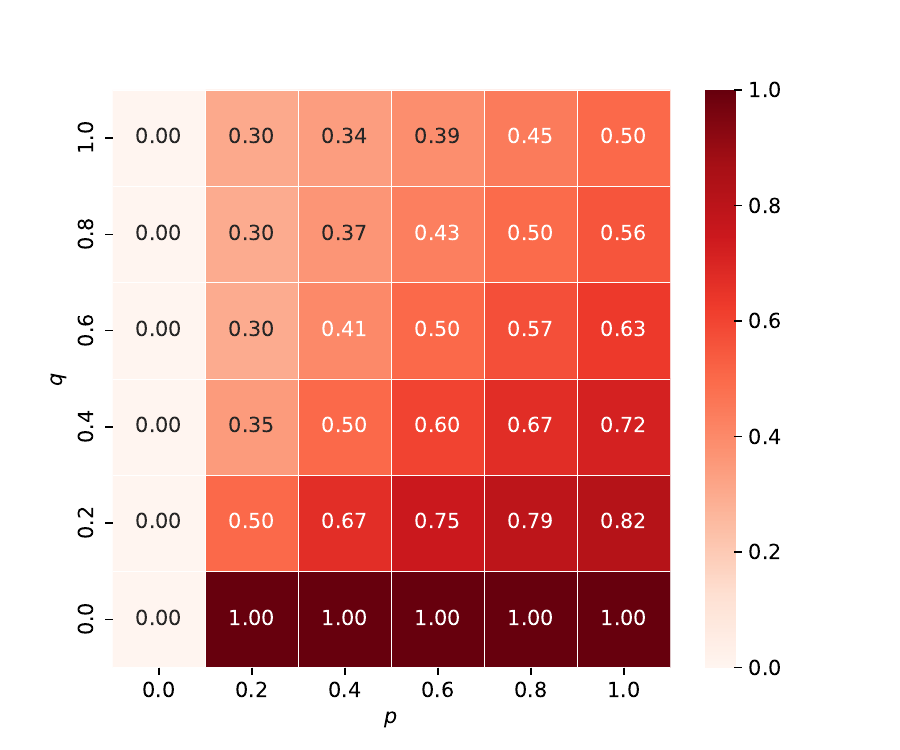}
        \label{fig:node_homophily}
    }
    \quad
    \subfigure{
        \includegraphics[width=0.4\textwidth]{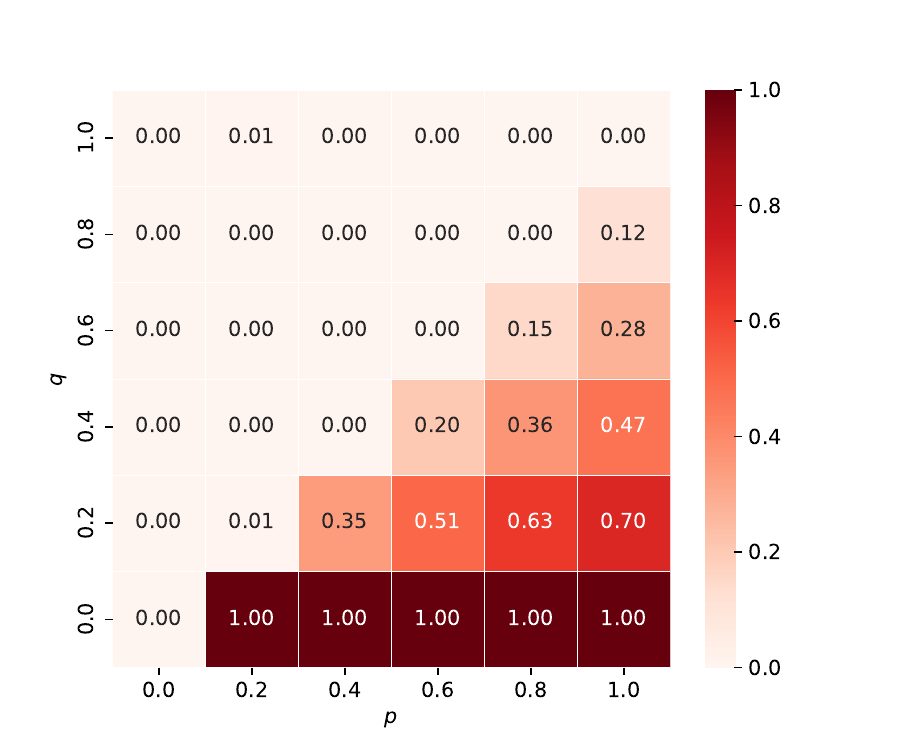}
        \label{fig:class_homophily}
    }
    \caption{The edge homophily (top-left), node homophily,(top-right), and class-insensitive homophily (bottom) ratios of the CSBM dataset.}
    \label{fig:CSBM Heatmaps}
\end{figure}

We provide additional homophily measures for the CSBM datasets measured with the edge homophily, node homophily, and class-insensitive homophily measure in Figure~\ref{fig:CSBM Heatmaps}.
We added a complete overview of CSBM results with different homophily measures in Table~\ref{tab:csbm summary}.

\begin{table}[ht]
\centering
\begin{tabular}{cccccccc}
\hline
Index & $p$ & $q$ & $\mathcal{H_E}$ & $\mathcal{H_N}$ & $\mathcal{H_{C}}$ & $h_{\text{adj}}$ & Accuracy \\
\hline
1  & 0.0 & 0.0 & 0.000 & 0.000 & 0.000 & 0.000 & 0.604 \\
2  & 0.0 & 0.2 & 0.000 & 0.000 & 0.000 & -1.000 & 0.604 \\
3  & 0.0 & 0.4 & 0.000 & 0.000 & 0.000 & -1.000 & 0.727 \\
4  & 0.0 & 0.6 & 0.000 & 0.000 & 0.000 & -1.000 & 0.708 \\
5  & 0.0 & 0.8 & 0.000 & 0.000 & 0.000 & -1.000 & 0.657 \\
6  & 0.0 & 1.0 & 0.000 & 0.000 & 0.000 & -1.000 & 0.700 \\
7  & 0.2 & 0.0 & 1.000 & 1.000 & 1.000 & 1.000 & 0.672 \\
8  & 0.2 & 0.2 & 0.500 & 0.500 & 0.005 & 0.0001 & 0.672 \\
9  & 0.2 & 0.4 & 0.333 & 0.349 & 0.000 & -0.333 & 0.693 \\
10 & 0.2 & 0.6 & 0.250 & 0.298 & 0.000 & -0.499 & 0.690 \\
11 & 0.2 & 0.8 & 0.200 & 0.298 & 0.000 & -0.600 & 0.654 \\
12 & 0.2 & 1.0 & 0.167 & 0.305 & 0.005 & -0.666 & 0.685 \\
13 & 0.4 & 0.0 & 1.000 & 1.000 & 1.000 & 1.000 & 0.721 \\
14 & 0.4 & 0.2 & 0.667 & 0.670 & 0.346 & 0.333 & 0.678 \\
15 & 0.4 & 0.4 & 0.500 & 0.500 & 0.001 & -0.000 & 0.699 \\
16 & 0.4 & 0.6 & 0.400 & 0.406 & 0.000 & -0.200 & 0.695 \\
17 & 0.4 & 0.8 & 0.333 & 0.367 & 0.000 & -0.333 & 0.691 \\
18 & 0.4 & 1.0 & 0.286 & 0.341 & 0.000 & -0.428 & 0.622 \\
19 & 0.6 & 0.0 & 1.000 & 1.000 & 1.000 & 1.000 & 0.720 \\
20 & 0.6 & 0.2 & 0.750 & 0.750 & 0.505 & 0.5000 & 0.624 \\
21 & 0.6 & 0.4 & 0.600 & 0.603 & 0.205 & 0.1999 & 0.779 \\
22 & 0.6 & 0.6 & 0.500 & 0.500 & 0.001 & -0.0005 & 0.623 \\
23 & 0.6 & 0.8 & 0.428 & 0.435 & 0.000 & -0.143 & 0.607 \\
24 & 0.6 & 1.0 & 0.375 & 0.388 & 0.000 & -0.250 & 0.657 \\
25 & 0.8 & 0.0 & 1.000 & 1.000 & 1.000 & 1.000 & 0.695 \\
26 & 0.8 & 0.2 & 0.800 & 0.794 & 0.633 & 0.600 & 0.654 \\
27 & 0.8 & 0.4 & 0.667 & 0.673 & 0.365 & 0.333 & 0.699 \\
28 & 0.8 & 0.6 & 0.571 & 0.574 & 0.152 & 0.142 & 0.697 \\
29 & 0.8 & 0.8 & 0.500 & 0.499 & 0.003 & -0.0005 & 0.630 \\
30 & 0.8 & 1.0 & 0.444 & 0.447 & 0.000 & -0.111 & 0.664 \\
31 & 1.0 & 0.0 & 1.000 & 1.000 & 1.000 & 1.000 & 0.694 \\
32 & 1.0 & 0.2 & 0.833 & 0.821 & 0.696 & 0.666 & 0.701 \\
33 & 1.0 & 0.4 & 0.714 & 0.718 & 0.472 & 0.427 & 0.754 \\
34 & 1.0 & 0.6 & 0.625 & 0.632 & 0.278 & 0.249 & 0.670 \\
35 & 1.0 & 0.8 & 0.556 & 0.557 & 0.120 & 0.110 & 0.665 \\
36 & 1.0 & 1.0 & 0.500 & 0.500 & 0.002 & -0.0002 & 0.689 \\
\hline
\end{tabular}
\caption{The results for the CSBM datasets with homophily ratios, the respective $p$ and $q$ values, and accuracy.}
\label{tab:csbm summary}
\end{table}

\end{document}